\documentclass[10pt,letterpaper]{article}
\usepackage[utf8]{inputenc}
\usepackage[top=0.85in,left=2.75in,footskip=0.75in]{geometry}

\usepackage{graphicx}
\usepackage{colortbl}
\usepackage{xcolor}
\usepackage{amsmath, amssymb}
\usepackage{parskip}
\usepackage{booktabs}
\usepackage{pdflscape}
\usepackage{multirow}
\usepackage{rotating}
\usepackage{adjustbox}
\usepackage{caption}
\usepackage{subcaption}
\usepackage{authblk}
\usepackage{chngcntr}
\usepackage{makecell}
\usepackage{wrapfig}

\usepackage{changepage}
\usepackage{textcomp,marvosym}
\usepackage{cite}
\usepackage{nameref,hyperref}
\usepackage[right]{lineno}
\usepackage[nopatch=eqnum]{microtype}
\DisableLigatures[f]{encoding = *, family = * }
\usepackage[table]{xcolor}
\usepackage{array}
\usepackage{float}

\newcolumntype{+}{!{\vrule width 2pt}}

\newlength\savedwidth

\raggedright
\usepackage[aboveskip=1pt,labelfont=bf,labelsep=period,justification=raggedright,singlelinecheck=off]{caption}

\makeatletter
\renewcommand{\@biblabel}[1]{\quad#1.}
\makeatother

\usepackage{lastpage,fancyhdr,graphicx}
\usepackage{epstopdf}
\fancyheadoffset[L]{2.25in}
\fancyfootoffset[L]{2.25in}
\begin{document}
\newgeometry{top=0.85in,left=1in,right=1in,footskip=0.75in}
\begin{flushleft}
{\Large
\textbf{Deep learning approaches show promise for predicting childhood malnutrition: a comparative study with traditional machine learning methods using survey data}
}
\newline
\\
Deepak Bastola\textsuperscript{1,2*},
Yang Li\textsuperscript{2},
\\
\bigskip
\textbf{1} Texas College of Management and IT, Kathmandu, Nepal\\
\textbf{2} Department of Mathematics and Statistics, Florida Atlantic University, Boca Raton, Florida, USA

\bigskip

* deepakbastola89@gmail.com

\end{flushleft}

\section*{Abstract}
Childhood malnutrition remains a major public health concern in Nepal and other low-resource settings, while conventional case-finding approaches are labor-intensive and frequently unavailable in remote areas. This study provides one of the first applications of machine learning and deep learning to identify child malnutrition in Nepal. We systematically compared 16 algorithms spanning deep learning, gradient boosting, and traditional machine learning families, using data from the Nepal Multiple Indicator Cluster Survey (MICS) 2019. A composite malnutrition indicator was constructed by integrating stunting, wasting, and underweight status, and model performance was evaluated using ten metrics, with emphasis on F1-score and recall to account for substantial class imbalance and the high cost of failing to detect malnourished children. Among all models, TabNet achieved the highest scores among evaluated models, likely attributable to its attention-based architecture. A consensus feature importance analysis identified maternal education, household wealth index, and child age as the primary predictors of malnutrition, followed by geographic characteristics, vaccination status, and meal frequency. Collectively, these results demonstrate a scalable, survey-based screening framework for identifying children at elevated risk of malnutrition and for guiding targeted nutritional interventions. The proposed approach supports Nepal’s progress toward the Sustainable Development Goals and offers a transferable methodological template for similar low-resource settings globally.

\bigskip
\textbf{Keywords:} childhood malnutrition, machine learning, deep learning, predictive modeling, Nepal



\section*{Introduction}

Malnutrition is a disorder that results from an imbalance between the body's nutritional needs and intake in the diet, and it encompasses both undernutrition and overnutrition. Malnutrition has many profound implications, including a weakened immune system, compromised growth and development of the brain, poorer educational performance, and increased risk of physical and mental diseases \cite{sideeffect_immune, sideeffect2, sideeffect3, halfdeath}. Malnutrition is responsible for almost half of all global deaths among children under the age of five \cite{half-child-deaths-linked-malnutrition}. According to the World Health Organization (WHO), Joint Child Malnutrition Estimates (JME) 2025 edition by UNICEF, and the World Bank, an estimated 150.2 million children under five (23.2\%) are stunted (too short for their age), while approximately 12.2 million suffer from severe wasting (too thin for their height) and 35.5 million children (5.5\%) are overweight \cite{uww}. It shows that there is still a long way to go in eliminating child malnutrition on a global scale even though substantial progress has been made over the past two decades. 

Every nation faces one or more forms of malnutrition, making its prevention and control one of today's most urgent global health challenges \cite{uww}.  Asia continues to be the global hotspot for child malnutrition, accounting for over half of all stunted children and nearly 70\% of all children affected by wasting under the age of five. The Nepal Demographic and Health Survey (NDHS) 2022 reported that in Nepal, around 25\% of children under five were stunted, 8\% were wasted, 19\% were underweight, and 1\% were overweight. In response, Nepal's government has established ambitious goals to curb these figures by 2030; reducing stunting to 15\%, wasting to 4\%, and underweight to 10\%. Child malnutrition in Nepal is rooted in a constellation of socioeconomic, geographic, maternal, and health-related factors \cite{indicators_nepal1}. Early detection of malnutrition and its risk factors is crucial for timely intervention and the prevention of associated health complications. 

This study presents a comprehensive evaluation of machine learning and deep learning architectures for predicting childhood malnutrition in Nepal using the Nepal Multiple Indicator Cluster Survey (NMICS) 2019 dataset. Given the non-linear separability of the data (Fig~\ref{fig8}), we employed multiple algorithms capable of capturing complex relationships among risk factors and malnutrition outcomes. The analysis incorporated variables spanning child care and supervision, health and nutrition indicators, child characteristics, maternal and household factors, and geographic determinants to address malnutrition's multifactorial etiology. We created a composite malnutrition indicator combining underweight, stunting, and wasting status, then systematically compared 16 models across three categories: deep learning (DNN, wide \& deep, ResNet, TabNet), gradient boosting (AdaBoost, CatBoost, XGBoost, LightGBM, histogram gradient boosting), and traditional machine learning (SVM, LDA, random forest, extra trees, decision tree, KNN, logistic regression). Model performance was evaluated using ten metrics: F1-score, recall, precision, accuracy, ROC-AUC, average precision, balanced accuracy, Cohen's kappa, Matthews correlation coefficient, and Brier score. Feature importance was determined through consensus ranking using mutual information scores, gradient boosting feature weights, random forest importances, and L1-regularized logistic regression coefficients. Deep learning methods, particularly TabNet, {showed marginal performance advantages over traditional ML methods}, achieving the highest F1-score, precision, and balanced accuracy, while AdaBoost and SVM led their respective categories with {the competitive} performance. These findings establish deep learning architectures, particularly those with attention mechanisms, as optimal for malnutrition prediction tasks in survey data.

Nepal has experienced persistent childhood malnutrition, a significant factor contributing to its classification among least developed countries \cite{sideeffect3, sideeffect2}. This study aims to identify the most effective predictive models and key determinants of under-five malnutrition in Nepal. While previous research has examined malnutrition determinants in Nepal \cite{indicators_nepal1, indicators_nepal2, indicators_nepal4} using classical statistical methods, no studies have applied machine learning or deep learning approaches to predict malnutrition outcomes in this context. International studies have explored machine learning for malnutrition prediction with varying methodological approaches (Table \ref{tab:previous_studies}). Research includes stunting prediction in Papua New Guinea \cite{papua_new_guinea}, comprehensive malnutrition assessment in Ethiopia \cite{ethiopia}, wasting identification in Egypt, and anthropometric indices in India \cite{india}. However, these investigations lack systematic comparison across deep learning, gradient boosting, and traditional machine learning categories, creating uncertainty about optimal methods across varying sample sizes and model complexities. This comprehensive evaluation addresses this gap by systematically comparing 16 models across three architectural categories, providing evidence-based guidance for predictive model selection in malnutrition surveillance. The findings can not only inform policy interventions in Nepal but also contribute for strategies in other nations facing similar challenges and speed up progress toward the sustainable development goals related to children's health.
\section*{Materials and Methods}

\subsection*{Data Set}

\subsubsection*{Data Source}
This cross-sectional study utilized data from the Nepal Multiple Indicator Cluster Survey (NMICS) 2019. It was conducted by the Central Bureau of Statistics (CBS) of Nepal with technical and financial assistance from UNICEF \cite{NMICS2019}. NMICS applied a stratified cluster sampling design using 512 clusters across seven provinces in Nepal to implement an extensive survey of 12,800 households. It ensures careful and consistent information on maternal and child health, nutrition, and related demographic indicators. In the data set, 6,749 children aged below five years were eligible for anthropometric measurement. After a data cleaning exercise in which missing and miscoded values were removed, 6,416 valid cases were preserved for data analysis.

\subsubsection*{Variables and Measurements}
\label{sec:var}

\textit{Response variable creation:}  We used three WHO-standardized anthropometric z-scores: weight-for-age (WAZ), height-for-age (HAZ), and weight-for-height (WHZ). Following WHO guidelines, children with z-scores below -2 standard deviations were classified as underweight, stunted, and wasted, respectively \cite{WHO2006}. A composite binary target variable `malnutrition' was created, where a child was classified as malnourished if they exhibited any form of undernutrition (underweight or stunted or wasted). The rationale for using a composite outcome was primarily driven by class imbalance considerations. In the data set,  1,520 children (23.69\%) were underweight, 2,077 children (32.37\%) were stunted, and 763 children (11.89\%) were wasted, individual conditions exhibited severe imbalance, with wasting alone being particularly problematic for classification modeling. Combining these into a composite outcome, the final data set $(n=6,416)$ has moderate class imbalance with 3,671 (57.2\%) nourished and 2,745 (42.8\%) malnourished children.

\begin{table}[!ht]
\centering

\begin{tabular}{lccc}
\toprule
\textbf{Conditions present} & \textbf{n} & \textbf{\% of total} & \textbf{\% of malnourished} \\
\midrule
\textit{No malnutrition} &
3,671 & 57.2\% & -- \\
\midrule
\multicolumn{4}{l}{\textit{Single condition}} \\
\quad Stunted only & 982 & 15.3\% & 35.8\% \\
\quad Wasted only & 243 & 3.8\% & 8.9\% \\
\quad Underweight only & 123 & 1.9\% & 4.5\% \\
\midrule
\multicolumn{4}{l}{\textit{Two conditions}} \\
\quad Stunted + Wasted & 0 & 0\% & 0\% \\
\quad Stunted + Underweight & 877 & 13.7\% & 31.9\% \\
\quad Wasted + Underweight & 302 & 4.7\% & 11.0\% \\
\midrule
\multicolumn{4}{l}{\textit{Three conditions}} \\
\quad All three (Stunted + Wasted + Underweight) & 218 & 3.4\% & 7.9\% \\
\midrule
\textit{Total malnourished} & {2,745} & 42.8\% & {100\%} \\
\bottomrule
\end{tabular}
\caption{Overlap structure of malnutrition conditions in the dataset$(n=6,416)$}
\label{tab:malnutrition_overlap}

\end{table}
Stunting was the most prevalent condition, while wasting was least common -- a pattern consistent with national epidemiological surveys in Nepal. The overlap structure (Table \ref{tab:malnutrition_overlap}) demonstrates substantial co-occurrence of conditions with 1,397 children (51\% of malnourished) presenting multiple concurrent indicators.

\textit{Domain-informed variable selection:} Based on established literature on child malnutrition determinants and expert consultation, we selected 23 available features across six thematic domains aligned with UNICEF's conceptual framework for malnutrition \cite{bangladesh1, papua_new_guinea, first, ethiopia, unicef2020conceptual}:
\begin{enumerate}
    \item \textit{Child care and supervision $(n=6)$}: days left alone for more than one hour, left with a child (under 10) for more than one hour, took away privileges, explained why the behavior was wrong, shook the child

    \item \textit{Child health and nutrition $(n=7)$}: ever breastfed, currently breastfed, meal frequency, diarrhea in the last two weeks, fever in the last two weeks, cough in the last two weeks, vaccination record

    \item \textit{Child characteristics $(n=4)$}: child age, child sex, child disability, child able to pick up a small object

    \item \textit{Maternal and household factors $(n=3)$}: maternal education level, household wealth index, health insurance coverage

    \item \textit{Geographic and environmental factors $(n=3)$:} residence type (urban/rural), provinces, safe stool disposal practices
\end{enumerate}

\begin{figure}[H]
    \centering
   \includegraphics[width=0.64\linewidth]{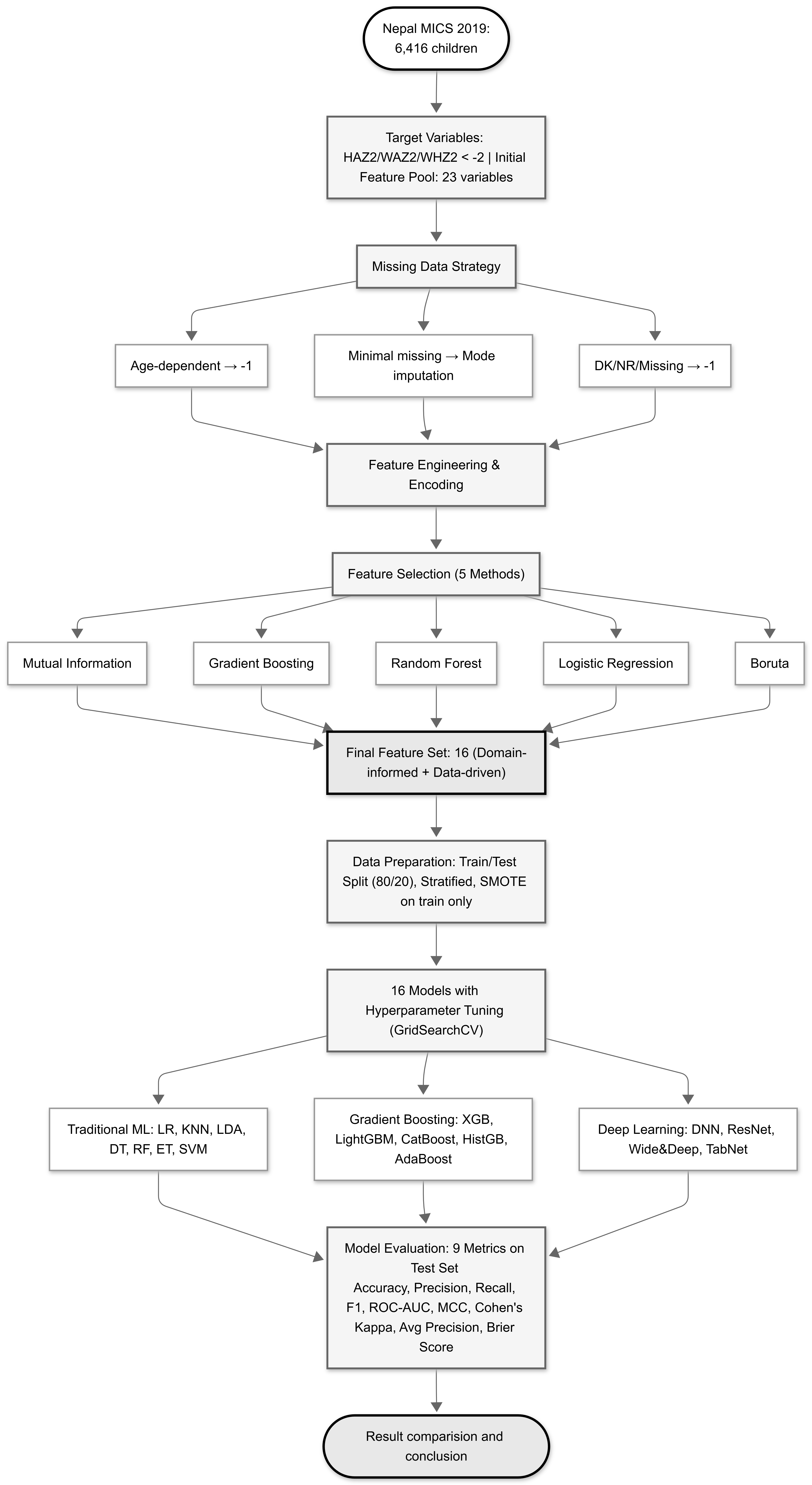}
    \caption{\bf Workflow diagram illustrating the malnutrition prediction
model development process}
    \label{fig1}
\end{figure}

To prevent circularity between outcome definition and predictors, we adopted two methodological safeguards: (1) anthropometric z-scores (HAZ, WHZ, WAZ) and their components (height, weight) were excluded from the predictor set, ensuring no direct mathematical components of the outcome appeared among input features; and (2) child age was operationalized as a categorical variable with five broad categories ($<1$, 1, 2, 3, 4 years) rather than continuous age in months. This categorical encoding captures developmentally meaningful stages of early childhood while deliberately avoiding fine-grained age information that could introduce circularity with HAZ and WAZ calculations, which require precise monthly age for WHO reference comparisons. The detailed workflow is presented in Fig~\ref{fig1}.

\subsubsection*{Pre-processing}
\label{sec:pre-processing}
We adopted a domain-informed approach to handle missing data, recognizing that the absence of responses often carries meaningful information in survey data. Age-dependent questions (e.g., breastfeeding status for older children, discipline practices) where ``not asked" occurred due to survey skip logic were encoded as -1, preserving this systematic missingness as an informative category. For variables with minimal missing data $(<1\%)$, such as recent illness indicators and health insurance status, mode imputation was applied. Response categories including ``don't know", ``no response", and ``missing/dk" were uniformly recoded as ``not asked" to maintain consistency across the dataset. This approach retained the maximum sample size while preventing information loss from survey design features. 

Features were encoded according to their measurement properties and theoretical relationships with malnutrition. Ordinal variables (wealth index: 1-5; mother's education: 0-3; meal frequency: 0-7) preserved their natural ordering. Age-dependent behavioral variables were ternary-encoded (1/0/-1 for yes/no/not asked) to distinguish genuine responses from non-applicability. Binary variables (sex, residence type, insurance) received standard 0/1 encoding. Multi-category nominal variables were transformed as follows: stool disposal practices were collapsed into binary safe/unsafe categories based on WHO guidelines; vaccination status indicated possession of any documentation; and province was one-hot encoded into six indicators, with Bagmati as the reference category. This encoding strategy balanced model interpretability with capturing non-linear relationships while avoiding multicollinearity.

To address the moderate class imbalance, we applied Synthetic Minority Over-sampling Technique (SMOTE)~\cite{chawla2002smote} within each cross-validation fold. SMOTE was applied exclusively to training data after the train-test split to prevent data leakage, generating synthetic minority class examples through interpolation. Test data retained the original distribution to ensure evaluation metrics reflected real-world model performance. The preprocessing pipeline was applied uniformly across all 16 machine learning and deep learning models to ensure fair comparison. Feature scaling using StandardScaler (zero mean, unit variance) was additionally applied to models sensitive to feature magnitudes: linear methods (Logistic Regression, LDA), distance-based methods (KNN, SVM), and all neural networks (DNN, ResNet, Wide \& Deep). Tree-based models (Decision Tree, Random Forest, Extra Trees), gradient boosting variants (XGBoost, LightGBM, CatBoost, Histogram Gradient Boosting, AdaBoost), and TabNet (which incorporates internal feature processing) used unscaled features, consistent with their algorithmic requirements. StandardScaler was fitted exclusively on training data and applied to test data to prevent data leakage. This standardized pipeline ensures that observed performance differences reflect genuine algorithmic capabilities rather than preprocessing inconsistencies.

\subsection*{Feature Selection}
We implemented a comprehensive multi-method feature selection strategy, integrating filter, wrapper, embedded, and other relevant approaches to identify the most predictive feature subset.

\subsubsection*{Methodology}
\textit{Filter methods:} Mutual information, Chi-square test, ANOVA F-statistic, Pearson correlation, and variance threshold to assess individual feature-target relationships.

\textit{Wrapper methods:} Recursive feature elimination (RFE) with logistic regression and gradient boosting, plus Sequential forward selection maximizing cross-validated F1-score.

\textit{Embedded methods:} Feature importance from random forest, gradient boosting, XGBoost, and L1-regularized logistic regression (LASSO).

\textit{Boruta algorithm:} All-relevant feature selection using random forest wrapper (100 iterations, $\alpha=0.05$) to identify features with statistically significant predictive power beyond random chance.

\textit{Ensemble aggregation:} Features were ranked by average position across all 10 methods, with confirmed Boruta features receiving priority weighting.

\subsubsection*{Final Features}
We selected 16 features based on Boruta confirmation supported by high ensemble rankings (average rank $\leq 14.3$) while capturing UNICEF's multi-dimensional malnutrition framework \cite{unicef2020conceptual}. This approach balanced statistical significance, domain relevance, and empirical performance (cross-validated $F1 = 0.6077 \pm 0.0836$). 

\textit{Selected features by category}\\
Child care and supervision (away privileges, left alone), child health and nutrition (vaccination record, meal frequency, recent diarrhea, recent cough), child characteristics (child age), maternal and household factors (mother's education, wealth index, health insurance), geographic and environmental factors (residence type, koshi, gandaki, karnali, sudurpaschim, safe stool disposal)

\begin{figure}[!h]
    \centering
   \includegraphics[width=0.85\linewidth]{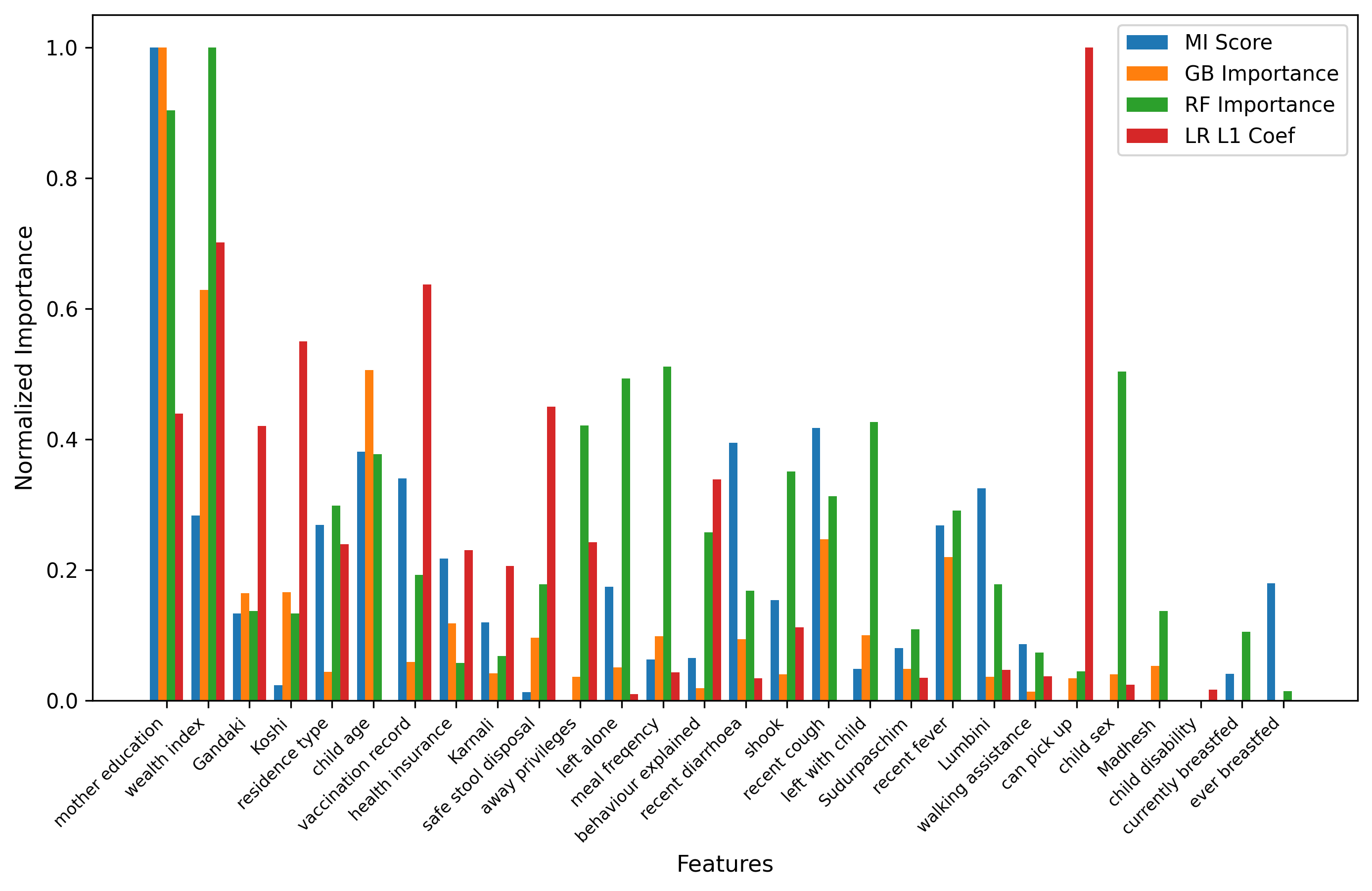}
    \caption{\bf Feature importance with their normalize score}
    \label{fig2}
\end{figure}

\begin{figure}[!ht]
    \centering
   \includegraphics[width=0.75\linewidth]{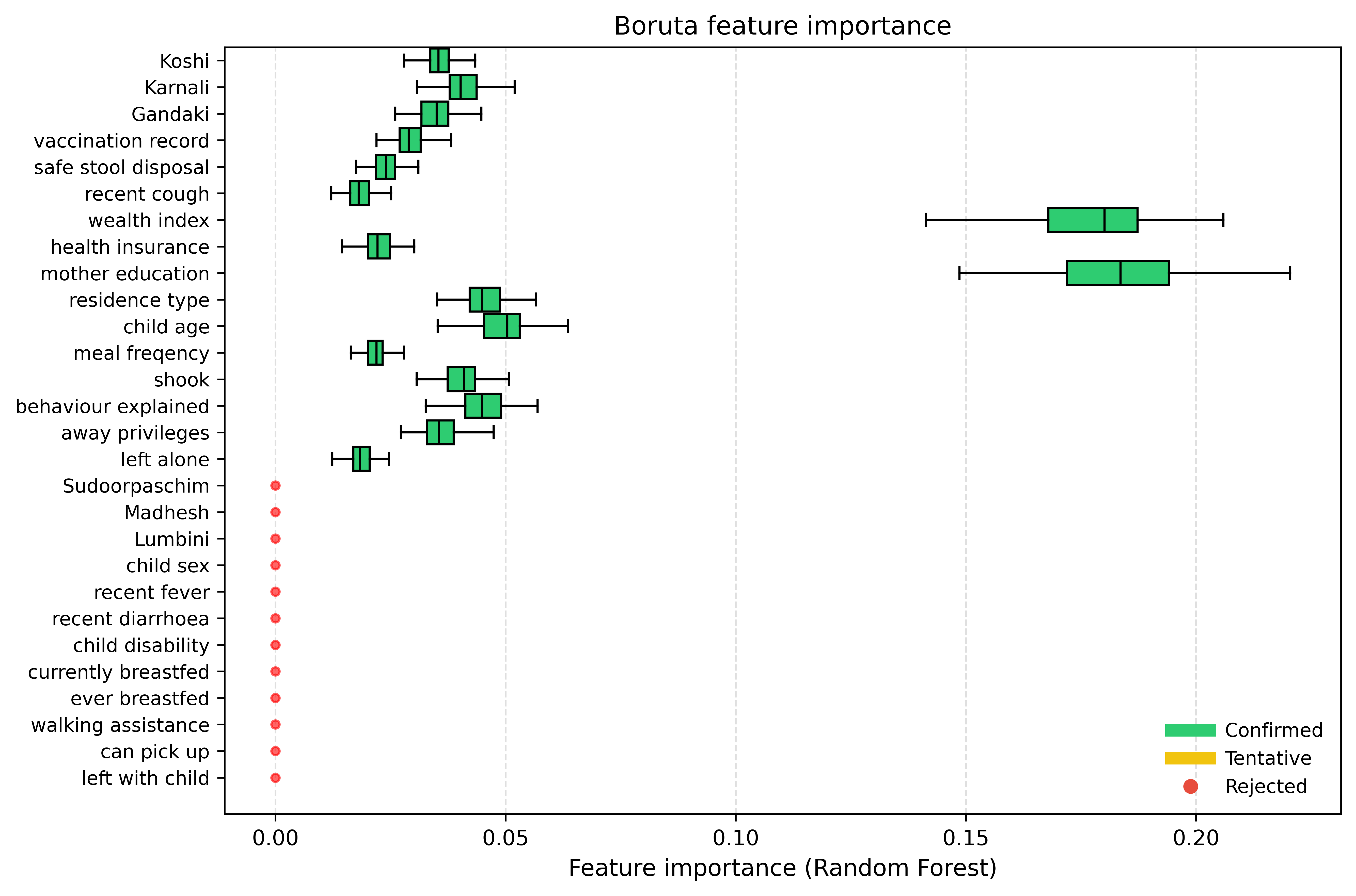}
    \caption{\bf Feature importance using Boruta method}
    \label{fig3}
\end{figure}

Features like recent diarrhea and sudurpaschim, though Boruta-rejected, were retained due to strong ensemble support and domain importance (WHO-recognized diarrhea-malnutrition pathway; geographic representation balance), as illustrated in Fig~\ref{fig2} and Fig~\ref{fig3}. The suitability of this multi-criteria feature selection strategy was empirically validated through a formal ablation analysis comparing our approach against strict Boruta-confirmed-only selection, with detailed results presented in Appendix~\ref{tab:ablation_analysis}. The ablation analysis confirmed that retaining these domain-relevant features improved model performance across all evaluation metrics compared to strict Boruta-only selection.


\subsection*{Methods}\label{sec:methods}
In this study, the response variable is binary, stating whether a child is malnourished or not. The aim is to make accurate predictions of malnutrition status for new cases that did not belong to the training set. Statistical procedures, machine learning models, or deep learning models can be used to solve this classification problem. All of these approaches aim to discover a function $f$ that takes input $\mathbf{x}$ to output $y$, based on observed input-output pairs $\{(\mathbf{x}_i, y_i)\}_{i=1}^n$. Model parameters are updated in an iterative process to minimize measures of error or maximize goodness of fit, such as the likelihood function. 

\subsubsection*{Traditional Machine Learning Models}
\textbf{Logistic regression}: Logistic regression is a widely-used statistical model for binary classification. It models the log-odds of the probability as a linear function of the predictors
\[
\log\left(\frac{\pi(\mathbf{x})}{1-\pi(\mathbf{x})} \right)= \mathbf{x}^\top \boldsymbol{\beta}.
\]
The parameter vector $\boldsymbol{\beta}$ can be estimated via maximum likelihood estimation, which then leads to the prediction of probability that an unseen case is malnourished.

\textbf{K-nearest neighbor}: K-nearest neighbor (KNN) is a non-parametric supervised machine learning algorithm that can be used for both classification and regression \cite{knn}. In the classification settings, KNN assigns the class that is most frequent among the $k$ nearest neighbors of the observation,
\[
\hat{y}(\mathbf{x}) = \arg\max_{c \in \mathcal{C}} \sum_{i \in \mathcal{N}_k(\mathbf{x})} \mathbf{1}\{y_i = c\}
\]
where $\mathcal{C}$ is the set of all possible classes, and $\mathcal{N}_k$ is the $k$ nearest neighbors of the observation with predictor $\mathbf{x}$. The assignment is done by majority vote.

\textbf{Decision tree}: A decision tree classifier (DT) is a non-parametric supervised learning algorithm that recursively divides the input space into disjoint regions $\{R_m\}_{m=1}^M$, each corresponding to a terminal node or leaf \cite{DT}. For any query point $\mathbf{x} \in \mathbb{R}^d$, the model assigns a predicted class label based on the majority class in the region $R_m$ that contains $\mathbf{x}$. This is formally represented as
$$
\hat{y}(\mathbf{x}) = \arg\max_{c \in \{0,1\}} \hat{p}_c^{(m)},
\quad \text{where } \mathbf{x} \in R_m,
$$
and the term $\hat{p}_c^{(m)}$ denotes the empirical proportion of class $c$ in region $R_m$, computed as
$$
\hat{p}_c^{(m)} = \frac{1}{|R_m|} \sum_{\mathbf{x}_j \in R_m} \mathbb{I}(y_j = c).
$$

\textbf{Random forest}: Random forest (RF) is an ensemble-based classification method that builds a collection of decision trees, each trained on a different bootstrap sample of the data and a random subset of features. This approach leverages the principle of aggregating the predictions of multiple uncorrelated models. During inference, each tree casts a ``vote" for the predicted class, and the final output is determined by majority voting across the ensemble. The randomness introduced through both data sampling and feature selection reduces the correlation between individual trees, thereby enhancing model stability and minimizing overfitting. As the number of trees increases, the model generally becomes more robust and accurate, especially in high-dimensional or noisy datasets \cite{rf}.

\textbf{Extremely randomized trees}: An ensemble learning method that constructs multiple decision trees using random subsets of features and random split points, rather than optimal splits. This additional randomness reduces variance more effectively than Random Forests while maintaining low bias, often improving generalization and reducing overfitting \cite{geurts2006extremely}.

\textbf{Linear discriminant analysis}: Linear discriminant analysis (LDA) is a supervised classification technique that projects data onto a lower-dimensional space to maximize class separation and minimize variance within classes. LDA is similar to analysis of variance and regression analysis, as it expresses one response variable as a linear combination of other features to find the best separator among classes \cite{LDA}.

\textbf{Support vector machine}: Support vector machine (SVM) is a widely used supervised machine learning method for classification and regression analysis. It constructs a hyperplane that maximizes the distance to the closest data points from any class. A larger margin generally leads to better generalization performance, reducing the likelihood of overfitting and improving the model's ability to perform well on unseen data. For linearly separable data, SVM solves the optimization problem
\[
\min_{w, b} \ \frac{1}{2} \|w\|^2 \quad \text{subject to} \quad y_i(w^\top x_i + b) \geq 1, \quad \forall i,
\]
where $x_i \in \mathbb{R}^d$ are input vectors and $y_i \in \{-1, +1\}$ are class labels.\\
In this study, we applied both a \textit{linear kernel} (hyperparameter-tuned model), which assumes data can be separated in the original feature space, and the \textit{Radial Basis Function kernel} (default model), a non-linear kernel that maps inputs into a higher-dimensional space using
$
K(x_i, x_j) = \exp\left(-\gamma \|x_i - x_j\|^2\right),
$
where $\gamma$ is a kernel parameter controlling the spread \cite{SVM}.

\subsubsection*{Gradient boosting models}

\textbf{Extreme gradient boosting}: Extreme gradient boosting (XGBoost) is an advanced ensemble learning method that has gained significant popularity due to its high predictive performance across a wide range of applications, including classification and regression tasks. The method is an extension of the gradient boosting framework, which constructs models in a sequential manner. At each stage, a new decision tree is trained with the specific purpose of addressing the errors or residuals left by the preceding trees. This iterative refinement enables the model to progressively improve accuracy and reduce bias. In the case of binary classification, XGBoost typically relies on a logistic loss function, which provides an effective way of quantifying classification errors \cite{xgb2}.

\textbf{Light gradient boosting machine}: A gradient boosting framework that employs histogram-based algorithms and Gradient-based One-Side Sampling (GOSS) with Exclusive Feature Bundling (EFB) to achieve faster training speed and higher efficiency. LightGBM uses leaf-wise tree growth strategy rather than level-wise, resulting in better accuracy with fewer iterations \cite{ke2017lightgbm}.

\textbf{Categorical boosting}: A gradient boosting algorithm specifically designed to handle categorical features efficiently using ordered target statistics and oblivious decision trees. CatBoost implements ordered boosting to avoid prediction shift and uses symmetric trees to reduce overfitting while maintaining high prediction speed \cite{prokhorenkova2018catboost}.

\textbf{Adaptive boosting}: An ensemble meta-algorithm that combines multiple weak learners (typically shallow decision trees) by iteratively adjusting sample weights based on classification errors. Misclassified samples receive higher weights in subsequent iterations, forcing the algorithm to focus on difficult cases, resulting in a strong classifier from weak ones \cite{freund1997decision}.

\textbf{Histogram-based gradient boosting}: A scikit-learn implementation of gradient boosting that uses histogram-based decision tree learning, binning continuous features into discrete bins. This approach significantly reduces memory consumption and computational cost while enabling efficient handling of missing values and support for monotonic constraints.

\subsubsection*{Deep Learning Models}

\textbf{Deep neural network}: A feedforward artificial neural network with multiple hidden layers between input and output layers, utilizing batch normalization, dropout regularization, and LeakyReLU activation functions to learn hierarchical feature representations. The deep architecture enables automatic feature extraction and non-linear pattern recognition in complex datasets \cite{goodfellow2016deep}.

\textbf{ResNet}: A deep neural network architecture that employs residual connections (skip connections) to address the vanishing gradient problem in very deep networks. These shortcut connections allow gradients to flow directly through the network, enabling training of substantially deeper models while maintaining or improving performance. ResNet-style models often perform well compared to other neural approaches and serve as a useful baseline in tabular deep learning research \cite{gorishniy2021revisiting}.

\textbf{Wide and deep network}: A hybrid neural network architecture that combines a wide linear model (for memorization of feature interactions) with a deep neural network (for generalization through feature learning). This dual-path structure enables both memorization of specific patterns and generalization to unseen data, particularly effective for recommendation systems and tabular data \cite{cheng2016wide}.

\textbf{TabNet}: TabNet is a deep learning architecture specifically tailored for tabular (structured) data, developed by Google Cloud AI. It introduces a novel approach by using sparse feature selection through sequential attention mechanisms. It allows the model to focus on the most informative features at each decision step. This adaptive feature selection process improves both learning efficiency and model interpretability. At each step, TabNet learns a feature mask that determines which subset of input variables should be emphasized, enabling it to reason selectively across multiple layers. The architecture supports local interpretability by showing how features influence individual predictions, and global interpretability by assessing the overall contribution of each feature across the dataset. This combination of performance and explainability makes TabNet particularly effective for structured data modeling tasks \cite{tabnet}.

\subsection*{Model Evaluation}
The performance of the classification models discussed in section \ref{sec:methods} was evaluated using standard metrics, including accuracy, precision, recall, F1-score, Cohen's kappa, area under the ROC curve (AUC), and average precision. All these metrics were derived from the confusion matrix with four categories of outcomes: true positives (TP), representing malnourished children correctly identified as malnourished; true negatives (TN), representing nourished children correctly identified as nourished; false positives (FP), representing nourished children incorrectly predicted as malnourished; and false negatives (FN), representing malnourished children incorrectly predicted as nourished. These metrics ensure a comprehensive assessment of the predictive capacity of the model and its ability to balance precision and recall to detect malnutrition.

\textit{Accuracy}: Accuracy measures the proportion of all predictions that the model classified correctly, both malnourished and nourished children.  
\[
\text{Accuracy} = \frac{TP + TN}{TP + TN + FP + FN}
\]
A high accuracy indicates that the model is correctly classifying the majority of children in both categories.

\textit{Precision}: Precision is the proportion of children predicted as malnourished who are actually malnourished.
\[
\text{Precision} = \frac{TP}{TP + FP}
\]
High precision means the model can avoid false alarms when identifying malnourished children.

\textit{Recall}: Recall (also called Sensitivity or True Positive Rate) is the probability that the model correctly identifies a child as malnourished, given that the child is actually malnourished.
\[
\text{Recall} = \frac{TP}{TP + FN}
\]
A high recall means the model effectively detects malnourished children. This is crucial in malnutrition detection so that at-risk individuals are not overlooked.
If we miss a malnourished child, they might not get the needed intervention. If we wrongly flag a healthy child, they might get extra check-ups, but that’s less harmful.

\textit{F-1 score}: F-1 score is the harmonic mean of \textit{precision} and \textit{recall}, providing a single measure that balances false positives and false negatives.  
\[
\text{F-1 Score} = 2 \cdot \frac{\text{Precision} \cdot \text{Recall}}{\text{Precision} + \text{Recall}} = \frac{2TP}{2TP + FP + FN}
\]
It takes into account both the ability to correctly identify malnourished children (recall) and to avoid misclassifying nourished children as malnourished (precision).

\textit{Cohen’s kappa}: Cohen's Kappa (\(\kappa\)) assesses the level of agreement between the model’s predictions and the actual class labels, accounting for chance agreement.  
\[
\kappa = \frac{p_o - p_e}{1 - p_e},
\]
where \( p_o \) is observed agreement and \( p_e \) is expected agreement by chance (the probability that the model and true labels agree randomly). A higher \(\kappa\) value indicates stronger agreement beyond chance.

\textit{Area under the ROC curve}: AUC evaluates the ability to distinguish between classes across all possible classification thresholds. It summarizes the trade-off between the true positive rate (recall) and the false positive rate.
An AUC score of 1.0 indicates perfect classification, while a score of 0.5 suggests the model performs no better 
than random guessing. Higher AUC values reflect better overall discrimination between nourished and malnourished children.

\textit{Average precision}: Average precision summarizes the precision-recall curve by computing the weighted mean of precisions at each threshold, using the increase in recall from the previous threshold as the weight.
\[
\text{AP} = \sum_n (R_n - R_{n-1}) \cdot P_n,
\]
where \( P_n \) is precision at the \(n\)-th threshold and \( R_n \) is recall at the \(n\)-th threshold. Higher AP indicates better performance in ranking malnourished cases higher across varying decision thresholds.

\textit{Matthews correlation coefficient (MCC)}: The Matthews Correlation Coefficient (MCC) is a balanced classification metric that accounts for all confusion matrix elements, producing values between -1 (complete disagreement) and +1 (perfect prediction). Unlike accuracy, the MCC remains reliable for imbalanced datasets, making it ideal for malnutrition prediction, where class distributions are often skewed

\textit{Brier score}: The Brier Score quantifies probabilistic prediction accuracy by measuring the mean squared difference between predicted probabilities and actual outcomes (range: 0-1, lower is better). It assesses both discrimination and calibration, ensuring that predicted probabilities reflect the true risk of malnutrition.

\section*{Results}

\subsection*{Descriptive Results}

\begin{figure}[!ht]
    \centering
     \includegraphics[width=0.75\linewidth]{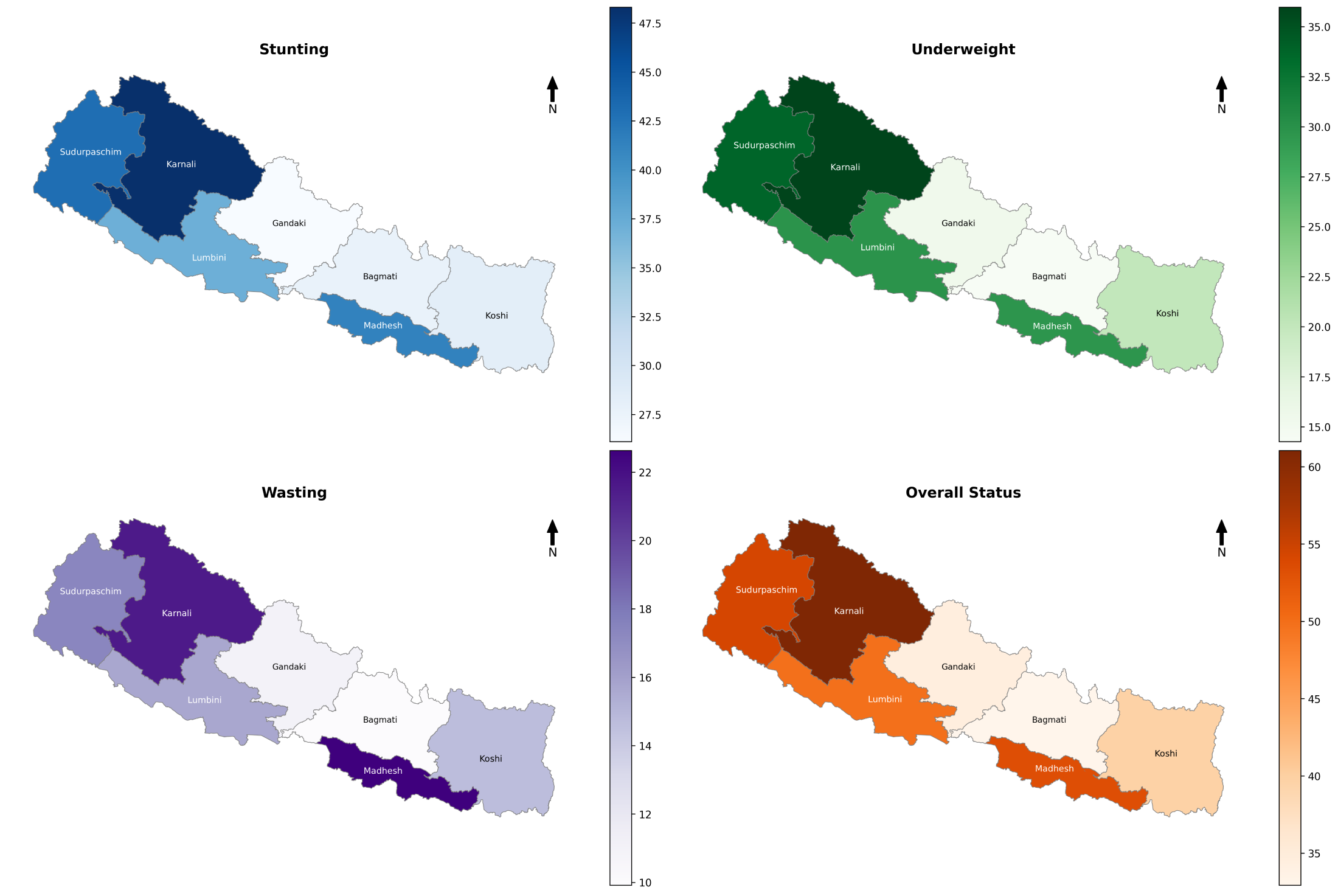}
    \caption{\textbf{Spatial distribution of child malnutrition prevalence across all provinces of Nepal.} Administrative boundary shapefiles obtained from 
geoBoundaries~\cite{runfola2020geoboundaries}; malnutrition prevalence values derived from Nepal NMICS 2019. Figure generated by the authors using Python (GeoPandas and Matplotlib)}
    \label{fig4}
\end{figure}

Fig~\ref{fig4} shows the spatial distribution of malnutrition across seven provinces in Nepal, grouped by stunting, underweight, wasting, and the combined malnutrition indicator. Karnali province has the highest percentage of stunting (48.31\%), underweight (35.97\%), and overall malnutrition (61.04\%), while Madhesh province has the highest wasting rate (22.63\%). Bagmati province and Gandaki province display relatively lower prevalence in all three indicators. These two provinces have a higher proportion of urban areas, and as indicated in Table~\ref{tab:descriptive_stat}, urban regions show lower malnutrition rates compared to rural areas.

\begin{table}[ !htbp]
\centering
\scriptsize
\setlength{\tabcolsep}{3pt}

\begin{minipage}{0.48\textwidth}
\centering
\begin{tabular}{lccc}
\hline
\multirow{2}{*}{\textbf{Variables}} & \multirow{2}{*}{\textbf{Total No. }} & \multicolumn{2}{c}{\textbf{Nutrition Status}} \\
\cline{3-4} & & \textbf{ Nourished n(\%)} & \textbf{ Malnourished n(\%)}  \\
&& -2 $\leq$ z-score $\leq$ 2 &  z-score $<$ -2 \\
& & n = 3671(\%) & n = 2745(\%) \\
\hline
\textbf{Mother education} & & & \\
None      & 1571 & 733 (46.7) & 838 (53.3) \\
Primary   & 2028 & 1101 (54.3) & 927 (45.7) \\
Secondary & 2337 & 1492 (63.8) & 845 (36.2) \\
Higher    & 480  & 345 (71.9) & 135 (28.1) \\
\hline

\textbf{Wealth index} & & & \\
Poorest & 1832 & 840 (45.9) & 992 (54.1) \\
Poorer  & 1317 & 748 (56.8) & 569 (43.2) \\
Middle  & 1275 & 749 (58.7) & 526 (41.3) \\
Richer  & 1170 & 728 (62.2) & 442 (37.8) \\
Richest & 822  & 606 (73.7) & 216 (26.3) \\
\hline

\textbf{Vaccination card} & & & \\
No  & 1012 & 524 (51.8)  & 488 (48.2) \\
Yes  & 2557 & 1572 (61.5) & 985 (38.5) \\
Not asked & 2847 & 1575 (55.3) & 1272 (44.7) \\
\hline

\textbf{Health insurance} & & & \\
No  & 6126 & 3465 (56.6) & 2661 (43.4) \\
Yes & 290  & 206 (71.0)  & 84 (29.0) \\
\hline

\textbf{Residence type} & & & \\
Urban & 2855 & 1503 (52.6) & 1352 (47.4) \\
Rural & 3561 & 2168 (60.9) & 1393 (39.1) \\
\hline

\textbf{Left alone ($>$ 1 hour)} & & & \\
Never  & 5020& 2945 (58.7) & 2075 (41.3) \\
One day  & 155 & 81 (52.3) & 74 (47.7) \\
Two days  & 253 & 141 (55.7) & 112 (44.3) \\
Three days  & 136 & 71 (52.2) & 65 (47.8) \\
Four days  & 116 & 68 (58.6) & 48 (41.4) \\
Five days  & 111 & 46 (41.4) & 65 (58.6) \\
Six days  & 57  & 33 (57.9) & 24 (42.1) \\
Seven days & 526 & 267 (50.8) & 259 (49.2) \\
Not asked & 42  & 19 (45.2) & 23 (54.8) \\
\hline

\textbf{Away privileges} & & & \\
No  & 3292 & 1854 (56.3) & 1438 (43.7) \\
Yes  & 2071 & 1096 (52.9) & 975 (47.1) \\
Not asked & 1053 & 721 (68.5) & 332 (31.5) \\
\hline

\textbf{Child age} & & & \\
$<1$ year & 1044 & 717 (68.7) & 327 (31.3) \\
One year & 1270 & 693 (54.6) & 577 (45.4) \\
Two years & 1259 & 688 (54.6) & 571 (45.4) \\
Three years & 1478 & 791 (53.5) & 687 (46.5) \\
Four years & 1365 & 782 (57.3) & 583 (42.7) \\
\hline

\textbf{Recent diarrhoea} & & & \\
No  & 5761 & 3335 (57.9) & 2426 (42.1) \\
Yes & 655  & 336 (51.3) & 319 (48.7) \\
\hline

\textbf{Gandaki province} & & & \\
No & 5705 & 3189 (55.9) & 2516 (44.1) \\
Yes & 711  & 482 (67.8)  & 229 (32.2) \\
\hline

\textbf{Koshi province} & & & \\
No & 5469 & 3069 (56.1) & 2400 (43.9) \\
Yes & 947  & 602 (63.6)  & 345 (36.4) \\
\hline

\textbf{Karnali province} & & & \\
No & 5673 & 3361 (59.2) & 2312 (40.8) \\
Yes & 743  & 310 (41.7)  & 433 (58.3) \\
\hline

\textbf{Meal frequency} & & & \\
No meals  & 529  & 359 (67.9)  & 170 (32.1) \\
One meal  & 93   & 53 (57.0)   & 40 (43.0) \\
Two meals  & 410  & 256 (62.4)  & 154 (37.6) \\
Three meals  & 527  & 292 (55.4)  & 235 (44.6) \\
Four meals  & 411  & 240 (58.4)  & 171 (41.6) \\
Five meals  & 200  & 111 (55.5)  & 89 (44.5) \\
Six meals  & 97   & 66 (68.0)   & 31 (32.0) \\
Seven meals  & 32   & 23 (71.9)   & 9 (28.1) \\
Not asked & 4117 & 2271 (55.2) & 1846 (44.8) \\
\hline

\textbf{Safe stool disposal} & & & \\
No  & 926  & 480 (51.8)  & 446 (48.2) \\
Yes  & 2644 & 1615 (61.1) & 1029 (38.9) \\
Not asked & 2846 & 1576 (55.4) & 1270 (44.6) \\
\hline

\textbf{Sudurpaschim prov} & & & \\
No & 5594 & 3276 (58.6) & 2318 (41.4) \\
Yes & 822  & 395 (48.1)  & 427 (51.9) \\
\hline

\textbf{Recent cough} & & & \\
No  & 5028 & 2863 (56.9) & 2165 (43.1) \\
Yes & 1388 & 808 (58.2)  & 580 (41.8) \\
\hline
\end{tabular}
\caption{Malnutrition by child and household characteristics}
\label{tab:descriptive_stat}
\end{minipage}
\hfill
\begin{minipage}{0.34\textwidth}
\small
Nepal has seven provinces: koshi, madhesh, bagmati, gandaki, lumbini, karnali, and sudurpaschim. As shown in Table \ref{tab:descriptive_stat}, the distribution of observations across provinces varies considerably. Among the provinces explicitly analyzed, Karnali province exhibits the highest malnutrition prevalence at 58.3\%, despite accounting for 11.58\% of the total sample. This is followed by Sudurpaschim province with 51.9\% malnutrition rate (12.81\% of sample). In contrast, Gandaki province shows the lowest malnutrition rate at 32.2\% (11.08\% of sample), while Koshi province demonstrates 36.4\% malnutrition (14.76\% of sample). The dataset captures a balanced representation of urban and rural areas, comprising 44.51\% and 55.49\% of the sample, respectively. Contrary to typical patterns, malnutrition is slightly lower among rural children (39.1\%) compared to urban children (47.4\%), suggesting potential urban poverty pockets or data-specific characteristics.

The age distribution of children is relatively balanced, with proportions ranging from 16.27\% to 23.04\% across five age groups. Notably, the youngest group (under 1 year) exhibits the lowest malnutrition rate at 31.3\%, while children aged one to three years show consistently higher malnutrition rates (45.4\% to 46.5\%). This pattern suggests increased vulnerability during the weaning period and early childhood development stages. There are pronounced differences in maternal education levels: only 7.48\% of mothers have higher education, and among their children, malnutrition is substantially lower (28.1\%). Conversely, children of mothers with no formal education, who constitute 24.49\% of the sample, experience the highest malnutrition rate (53.3\%). Children of mothers with secondary education (36.43\% of sample) show 36.2\% malnutrition, demonstrating a clear inverse relationship between maternal education and child nutritional status.

Household economic status, measured by the wealth index, reveals stark disparities across categories. The poorest quintile accounts for 28.56\% of the sample and exhibits the highest malnutrition rate at 54.1\%. This rate progressively decreases with increasing wealth: poorer (43.2\%), middle (41.3\%), richer (37.8\%), and richest (26.3\%). 
\end{minipage}
\end{table}

Children from the wealthiest families, representing 12.81\% of the sample, are substantially less likely to be malnourished compared to those from lower-income households. Health insurance coverage among children is minimal at only 4.52\%. Interestingly, children with health insurance demonstrate lower malnutrition rates (29.0\%) compared to uninsured children (43.4\%), suggesting that insurance coverage may serve as a proxy for better overall socioeconomic conditions and health-seeking behavior.

Health conditions reveal important vulnerabilities: 10.21\% of children experienced diarrhea recently, with those affected showing higher malnutrition rates (48.7\%) compared to those without diarrhea (42.1\%). Recent cough, reported in 21.64\% of children, shows a slightly lower association with malnutrition (41.8\% vs 43.1\%). Vaccination coverage indicates that 39.86\% of children have vaccination cards, and these children exhibit lower malnutrition (38.5\%) compared to those without cards (48.2\%). Safe stool disposal practices, recorded for 41.20\% of children, are associated with lower malnutrition rates (38.9\%) compared to unsafe disposal (48.2\%), highlighting the importance of sanitary practices in child nutrition outcomes.
Child care patterns reveal concerning trends: while 78.25\% of children are never left alone for more than one hour, those left alone five days per week show the highest malnutrition rate at 58.6\%. Meal frequency data, available for 35.82\% of children, show variation across categories, with children receiving no meals paradoxically showing lower malnutrition (32.1\%), likely reflecting data collection issues or proxy caregiving arrangements. 

\subsection*{Performance Evaluation}
Performance evaluation was conducted to enable systematic comparison across families of machine learning and deep learning models. Although SMOTE oversampling was applied to training data (Section~\ref{sec:pre-processing}), the test set retained the original imbalance class distribution, necessitating imbalance-aware evaluation. We prioritize F1-score and recall due to the cost asymmetry of false negatives and complement these with eight additional metrics (precision, accuracy, ROC-AUC, average precision, balanced accuracy, Cohen's kappa, MCC, Brier score) for comprehensive assessment. Table \ref{tab:model_performance} presents the comparative performance of machine learning and deep learning models across ten evaluation metrics.

TabNet emerges as the top-performing model, achieving the highest accuracy (0.62), precision (0.63), F1-score (0.62), ROC-AUC (0.64), balanced accuracy (0.62), Cohen's kappa (0.24), and MCC (0.24). Among traditional machine learning approaches, Support vector machine demonstrates competitive performance with a recall of 0.61 and ROC-AUC of 0.64. Linear discriminant analysis and logistic regression also show strong discriminative ability, both achieving ROC-AUC scores of 0.64 and average precision of 0.56. Within the gradient boosting family, AdaBoost yields the best results, particularly in average precision (0.56) and ROC-AUC (0.64), while XGBoost and LightGBM produce comparatively modest performance. Deep learning architectures DNN, Wide \& Deep, and ResNet -- exhibit consistently perform with metrics clustered between 0.60 and 0.62 across most measures. K-nearest neighbors records the lowest performance across the majority of metrics, suggesting limited suitability for this classification task.

\begin{table}[!ht]
\centering
\Large
\resizebox{\textwidth}{!}{
\begin{tabular}{l c c c c c c c c c c}
\toprule
\hline
\textbf{Model} & 
\textbf{Accuracy} & 
\textbf{Precision} & 
\textbf{Recall} & 
\textbf{F1 score} &
\textbf{ROC-AUC} & 
\makecell{\textbf{Average} \\ \textbf{Precision}} &
\makecell{\textbf{Balanced} \\ \textbf{Accuracy}} &
\makecell{\textbf{Cohen's} \\ \textbf{Kappa}} &
\textbf{MCC} &
\textbf{Brier} \\
\midrule
\hline

\multicolumn{9}{l}{\textbf{Deep Learning}} \\
Deep Neural Networks
& 0.61  &    0.60    & 0.61  &     0.60     & 0.62    &        0.53       &    0.59    &      0.18  & 0.19 &   0.24 \\
Wide \& Deep Networks 
& \textbf{0.62}   &    0.61 &   \textbf{0.62} &     0.61    & \textbf{0.64}     &      0.54 &         0.60      &   0.20 &0.20  & \textbf{0.23}\\
ResNet
& 0.61  &     0.60  &  0.61   &   0.60   &  0.63   &        0.54      &    0.59   &      0.19 & 0.19 &  \textbf{0.23} \\
TabNet 
& \textbf{0.62} & \textbf{0.63} & \textbf{0.62} & \textbf{0.62} & \textbf{0.64} & 0.54 & \textbf{0.62} & \textbf{0.24} & \textbf{0.24} & 0.24 \\

\hline
\multicolumn{9}{l}{\textbf{Gradient Boosting}} \\
AdaBoost 
& 0.60 & 0.61 & 0.60 & 0.61 & \textbf{0.64} & \textbf{0.56}  &        0.60     &    0.20 &0.20  & 0.24 \\
CatBoost 
& 0.60 & 0.60 & 0.60 & 0.60 & 0.63 & 0.54    &      0.59     &    0.19& 0.19  & 0.24 \\
Hist Gradient Boosting 
& 0.60 & 0.60 & 0.60 & 0.60 & 0.62 & 0.52  &        0.59     &    0.18 &0.18  & 0.24 \\
LightGBM 
& 0.60 & 0.60 & 0.60 & 0.60 & 0.62 &  0.53  &        0.59 &        0.18 &0.18  & 0.24 \\
XGBoost 
& 0.59 & 0.59 & 0.59 & 0.59 & 0.62 & 0.53    &      0.58  &       0.16& 0.16 &  0.24 \\

\hline
\multicolumn{9}{l}{\textbf{Traditional Machine Learning}} \\
Support Vector Machine 
& 0.61 & 0.62 & 0.61 & 0.61 & \textbf{0.64} & 0.54    &      0.61    &     0.21& 0.21  & 0.24 \\
Linear Discreminant Analysis 
& 0.60 & 0.61 & 0.60 & 0.60 & \textbf{0.64} & \textbf{0.56}   &       0.61      &   0.21& 0.21 &  0.24 \\
Extra Trees 
& 0.60 & 0.60 & 0.60 & 0.60 & 0.63 & 0.54     &     0.60     &    0.19& 0.19 &  0.24\\
Random Forest 
& 0.59   &    0.59  &  0.59  &    0.59   &  0.62    &       0.53 &         0.59     &    0.17 &0.17   &0.24 \\
Decision Tree 
&  0.58    &   0.62 &   0.58    &  0.58 &    0.61   &        0.51   &       0.60    &     0.19& 0.20  & 0.26 \\
KNN 
& 0.58 & 0.58 & 0.58 & 0.58 & 0.60 & 0.50     &     0.58  &       0.15 &0.15&   0.27 \\
Logistic Regression 
&  0.60    &   0.61  &  0.60  &    0.60  &  \textbf{0.64}    &       \textbf{0.56} &         0.60        & 0.20 &0.20   &0.24 \\
\hline
\bottomrule
\end{tabular}
}
\caption{Model performance}
\label{tab:model_performance}
\end{table}

\begin{figure}[!ht]
    \centering
     \includegraphics[width=0.85\linewidth]{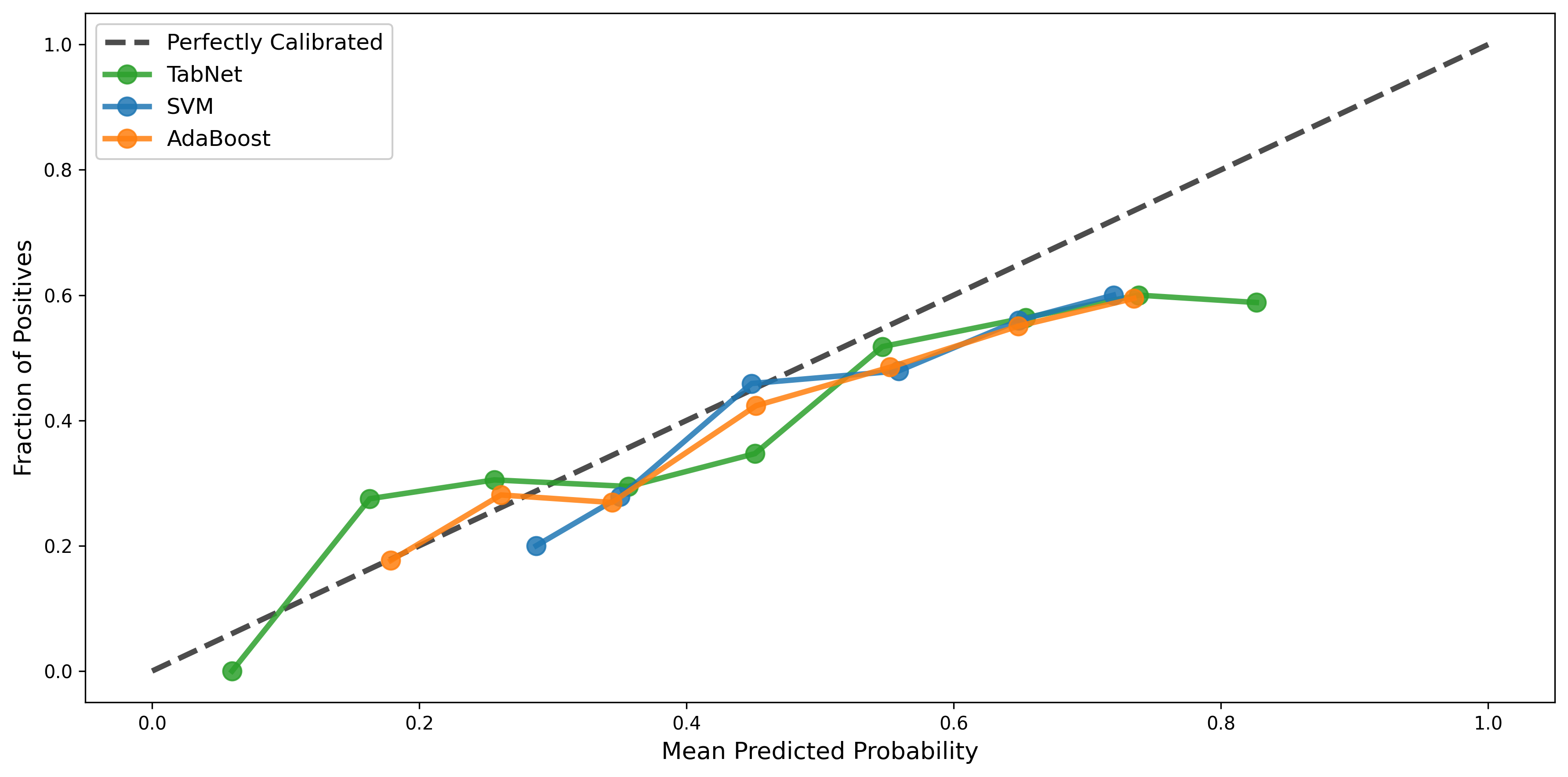}
    \caption{\bf Calibration curves of the top-performing model from each model category}
    \label{fig5}
\end{figure}

Fig~\ref{fig5} shows the calibration of malnutrition prediction, evaluating how well the predicted probabilities from the three best-performing models in each group (TabNet, SVM, and AdaBoost) correspond to the actual outcomes. The diagonal line indicates perfect concordance between predicted probabilities and the observed prevalence of malnutrition. Across most probability ranges, all three models show reasonable alignment with the ideal calibration line, indicating that predicted malnutrition risks generally correspond to observed outcomes. Some deviations are observed at lower and higher probability bins, reflecting under- or over-estimation of malnutrition risk in certain ranges. Well-calibrated models are essential in this context, as healthcare workers rely on accurate probability predictions to prioritize nutritional interventions and allocate limited resources effectively in resource-constrained settings like Nepal. 

\begin{figure}[!ht]
    \centering
     \includegraphics[width=0.85\linewidth]{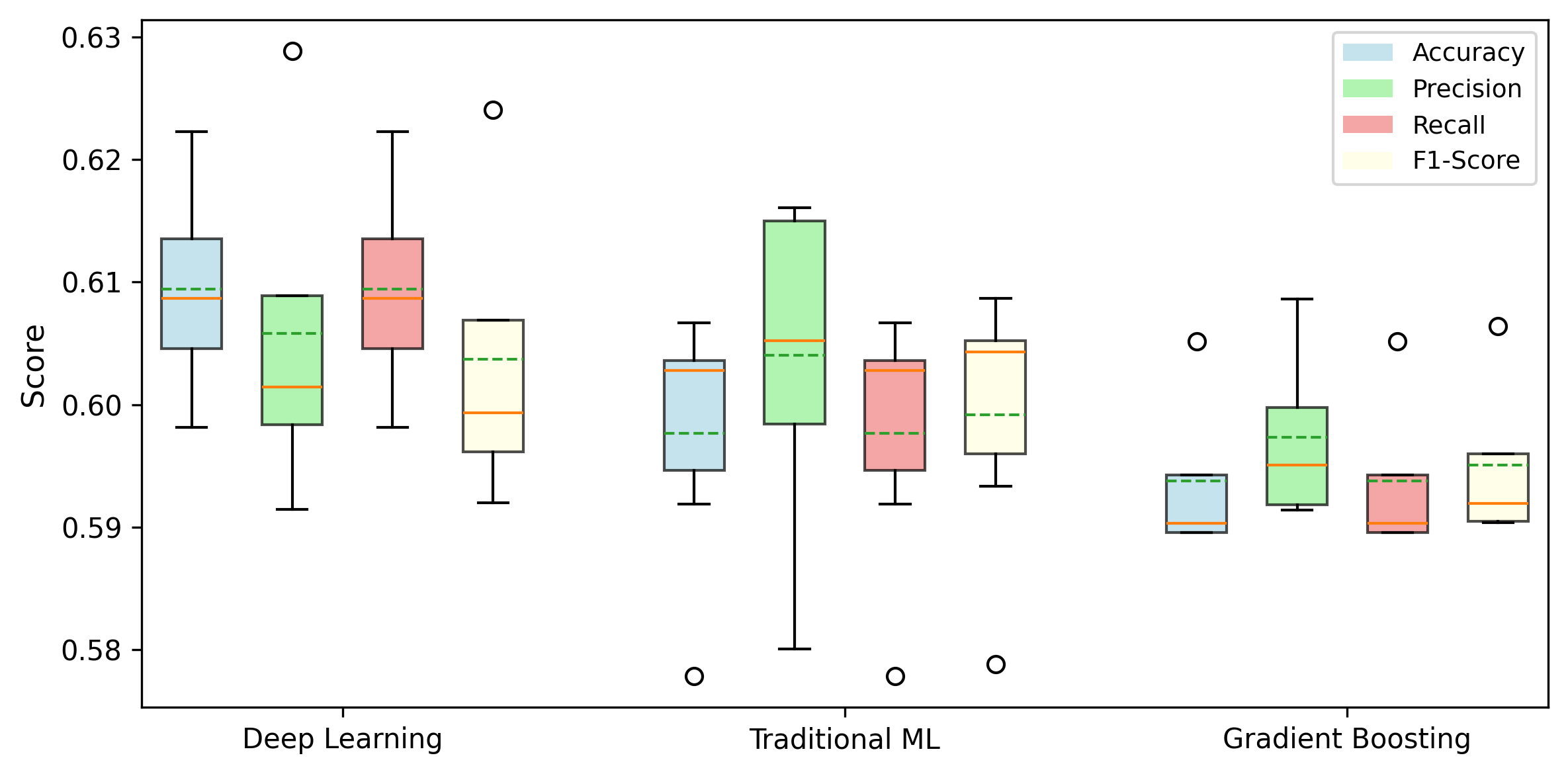}
    \caption{\bf Within-category performance variance}
    \label{fig6}
\end{figure}

Fig~\ref{fig6} shows performance metric distributions for three model categories: deep learning, traditional machine learning, and gradient boosting. Deep learning models have the highest median performance across accuracy, precision, recall, and F1-score, all near 0.61, with moderate variance in precision. Traditional machine learning models are more heterogeneous, especially in accuracy and recall (IQR $\approx 0.588-0.602$); precision has the widest spread, with a median around 0.606 and some models reaching 0.614. Gradient boosting models are the most consistent, with narrow interquartile ranges, few outliers, and all four metrics tightly clustered around 0.60.

\begin{figure}[!ht]
    \centering
     \includegraphics[width=0.99\linewidth]{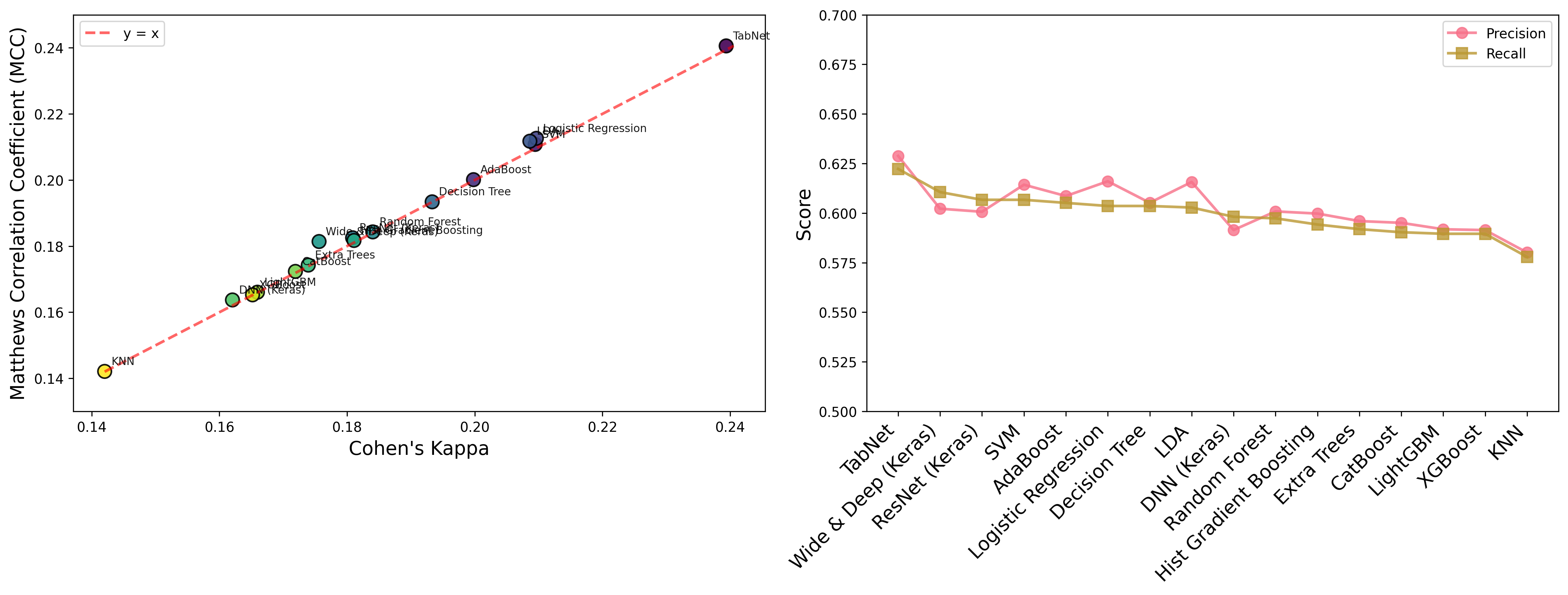}
    \caption{\bf Agreement metrics comparison and precision-recall trade-off}
    \label{fig7}
\end{figure}

Fig~\ref{fig7} shows two views of model performance. In the left panel, Cohen’s kappa and MCC both quantify agreement beyond chance. TabNet and SVM lie in the upper-right corner, with the strongest agreement, while KNN has the weakest (kappa: 0.15, MCC: 0.15). Most models fall along the $y = x$ diagonal line, indicating close correspondence between the two metrics, except Decision Tree, which has a notably higher MCC than kappa. The right panel shows precision–recall trade-offs. For most models, precision and recall are similar and range from 0.58 to 0.62, indicating balanced detection of true positives and control of false positives. TabNet attains the highest precision (0.63) with comparable recall (0.62). Deep learning models (leftmost) tend to have slightly higher precision, whereas gradient boosting methods (rightmost) show modest declines in both metrics. Decision Tree shows a sharp precision spike with lower recall, indicating overly conservative predictions.

\section*{Discussion}

This is, to our knowledge, one of the first use of machine and deep learning to predict under-five malnutrition in Nepal. Using the NMICS 2019 dataset, the study identified key determinants of childhood malnutrition. Maternal education was among the strongest predictors, consistent with evidence that maternal literacy improves child nutrition and healthcare use. Household wealth status was also strongly associated with malnutrition, reflecting inequalities in food security and healthcare access. Child age was important, likely capturing vulnerability during the complementary feeding period. Geographic region showed substantial predictive value, indicating regional differences in food availability, cultural feeding practices, and health services. Vaccination status was protective, probably as a marker of broader engagement with preventive healthcare rather than a direct causal factor. Children left unsupervised for at least one hour per day had higher malnutrition prevalence, plausibly due to irregular or inadequate feeding in the caregiver’s absence. 
Child age and recent illness (fever, diarrhea) also emerged as key predictors, consistent with anemia prediction using NDHS data \cite{bastola2026predicting} and suggesting shared etiological pathways across different childhood undernutrition outcomes in Nepal. 
Overall, these findings underscore the complex, multidimensional nature of childhood malnutrition spanning socioeconomic status, education, healthcare access, and caregiving patterns that must be captured by machine learning models for robust prediction.

Among deep learning architectures, TabNet exhibited consistently strong performance, followed by Wide \& Deep networks (F1: 0.61, recall: 0.62) and standard DNN and ResNet (both F1: 0.60, recall: 0.61). TabNet's advantage stems from its sequential attention mechanism and sparse feature selection, which explicitly learns which features to use at each decision step, particularly beneficial for tabular data with heterogeneous features like MICS surveys. Wide \& Deep networks combine memorization of specific feature interactions (wide component) with generalization through deep embeddings (deep component), enabling effective capture of both simple and complex patterns. However, standard DNNs and ResNet, while powerful for image and sequential data, lack specialized mechanisms for tabular feature relationships, explaining their marginally lower performance despite similar architectural depth.
Within gradient boosting methods, AdaBoost achieved the strongest performance (F1: 0.61, recall: 0.60, average precision: 0.56), outperforming CatBoost, LightGBM, histogram gradient boosting, and XGBoost (all F1: 0.60 or below). AdaBoost's success likely derives from its adaptive reweighting strategy, which iteratively focuses on misclassified instances, particularly valuable in imbalanced datasets where minority class examples (malnourished children) require greater attention. CatBoost, LightGBM, and histogram gradient boosting demonstrated nearly identical performance (F1: 0.60). XGBoost's relative underperformance (F1: 0.59, accuracy: 0.59) compared to other gradient boosting methods warrants careful consideration. While XGBoost employs sophisticated regularization techniques and parallel tree construction, its level-wise tree growth may be less effective than LightGBM's leaf-wise approach for datasets with subtle, localized patterns. Additionally, XGBoost's default hyperparameters are often tuned for larger datasets; the NMICS 2019 sample size may not fully leverage XGBoost's capacity, leading to either underfitting or suboptimal tree structures. The similar performance across LightGBM, CatBoost, and histogram gradient boosting (all achieving F1: 0.60) suggests that fundamental boosting principles matter more than specific algorithmic innovations for this classification task, and that hyperparameter tuning had a greater impact than architectural differences among these modern boosting variants.
Among traditional machine learning models, support vector machine emerged as the strongest performer (F1: 0.61, recall: 0.61, ROC-AUC: 0.64), nearly matching TabNet's recall and achieving the highest ROC-AUC alongside AdaBoost and linear discriminant analysis. SVM's success can be attributed to its ability to find optimal decision boundaries in high-dimensional spaces using kernel functions, effectively capturing non-linear relationships between sociodemographic predictors. Linear discriminant analysis also performed competitively (F1: 0.60, ROC-AUC: 0.64, average precision: 0.56). Logistic regression achieved similar performance (F1: 0.60, ROC-AUC: 0.64), demonstrating that well-calibrated linear models remain competitive for interpretable prediction tasks. Tree-based traditional methods showed mixed results. Extra trees and random forest achieved F1 scores of 0.60 and 0.59 respectively, with random forest slightly underperforming despite its reputation for robustness. This suggests that the additional randomness in extra trees' split selection provided marginal benefits for this dataset's feature space. Decision Tree (F1: 0.58) suffered from instability and overfitting to training data, as evidenced by its high precision (0.62) but lower recall (0.58), indicating overly conservative predictions. K-nearest neighbors recorded the poorest performance (F1: 0.58, ROC-AUC: 0.60), likely due to its sensitivity to feature scaling and the curse of dimensionality, where distance metrics become less meaningful.

Across model categories (Fig~\ref{fig6} and Fig~\ref{fig9}), deep learning achieved slightly higher median scores but with greater variance, traditional machine learning showed the widest range (KNN 0.58 to SVM 0.61), and gradient boosting produced the most stable but conservative results. This suggests that deep learning architectures like TabNet better capture complex patterns, traditional models like SVM remain highly competitive, and gradient boosting provides a reliable low-variance baseline.
The strong concordance between Cohen's kappa and MCC for all models (Fig~\ref{fig7}, left) confirms consistent performance rankings across agreement metrics. The small precision–recall gap (Fig~\ref{fig7}, right) shows that most algorithms balanced recall and precision, critical for both identifying malnourished children and avoiding unnecessary interventions.

Our initial candidate features were informed by established work in similar contexts, including studies from Bangladesh, India, Ethiopia, Philippines, and Papua New Guinea \cite{bangladesh1, first, ethiopia2, philippines, papua_new_guinea}. After applying our multi-criteria feature selection pipeline (detailed in section ~\ref{sec:var}), the final feature set included commonly identified determinants across countries was child age, wealth index, maternal education, and geographic region. However, novel features specific to our dataset such as away privileges and left alone, which captured child supervision patterns not consistently examined in prior work. This shared set of features indicates the presence of common underlying etiological pathways for childhood malnutrition, whereas the context-specific features primarily capture variation arising from differences in information availability across datasets.

\noindent
\begin{minipage}[t]{0.55\textwidth}
\adjustbox{valign=t}{%
\scalebox{0.6}{ %
\rotatebox{90}{%
\renewcommand{\arraystretch}{1.4}
\rowcolors{3}{gray!10}{white}
\begin{tabular}{|c|m{3.6cm}|c|c|c|c|*{5}{c|}*{7}{c|}}
\hline
\rowcolor{gray!30}
\multirow{3}{*}{Author} & 
\multirow{3}{*}{Model} & 
\multirow{3}{*}{Country} & 
\multirow{3}{*}{Year} & 
\multirow{3}{*}{Data Source} & 
\multirow{3}{*}{Size} & 
\multicolumn{5}{c|}{\textbf{Anthropometric}} & 
\multicolumn{7}{c|}{\textbf{Performance Comparison Metric}} \\ \cline{7-18}

\rowcolor{gray!30}
Authors & Employed Methods & Country & Year & Data Source & Size & 
Stunting & Wasting & Underweight & \begin{tabular}{c}Stunted-\\Wasted \end{tabular} & \begin{tabular}[c]{@{}c@{}}Stunted-Wasted-\\Underweight\end{tabular} & 
Accuracy & Precision & Recall & \begin{tabular}{c}
F1- \\ Score \end{tabular}& AP & AUC & \begin{tabular}{c} Cohen's \\ Kappa \end{tabular}\\ 
\hline

\begin{tabular}{c}
     Bastola \& Li  \\
     (this study)
\end{tabular} & \begin{tabular}[c]{@{}m{3.9cm}@{}}DNN, Wide\&Deep,LDA, \\ResNet, \textbf{TabNet}, KNN,\\ \textbf{AdaBoost}, CatBoost,\\HGBoost,LGBM,LR,RF,\\ ET,XGBoost,\textbf{SVM}, DT\end{tabular} 
& Nepal & 2025 & NMICS 2019 & 6,416 & 
N/A & N/A & N/A & N/A & \checkmark & 
\checkmark & \checkmark & \checkmark & \checkmark & \checkmark & \checkmark & \checkmark \\
\hline

 Tamanna et al.~\cite{first} & \begin{tabular}{c} LR, KNN, NN, DT,\\ XGBoost, SVM, \textbf{RF} \end{tabular} 
& Bangladesh & 2025 & BDHS 2022 & 7,910 & 
\checkmark & \checkmark & \checkmark & N/A & N/A & 
\checkmark & \checkmark & \checkmark & \checkmark & N/A & N/A & \checkmark \\
\hline
 Hendy et al.~\cite{egypt} &	 LR, RF, KNN, DT, \textbf{XGBoost}, Gradient Boosting &	Egypt	& 2024 &	\begin{tabular}{c}
     EDHS 2005,  \\ 2008 \& 2014 \end{tabular}  &		35,720 & N/A &	\checkmark	& N/A	& N/A &	N/A &	\checkmark	& \checkmark &	\checkmark &	\checkmark &	N/A &	\checkmark &	N/A\\
\hline
	Islam et al.~\cite{bangladesh2}	& LR, \textbf{XGBoost}, ANN, RF	& Bangladesh	&2024&	BDHS 2017-18&	 7,777 & \checkmark	& \checkmark	& \checkmark	 & N/A	&N/A	& \checkmark	& \checkmark	& \checkmark	& \checkmark&	N/A	& \checkmark	&N/A \\
\hline
Shen et al.~\cite{papua_new_guinea} & \textbf{XGBoost}, LR, DT,  SVM	& \begin{tabular}{c} Papua New  \\   Guinea \end{tabular} 	&2023&	\begin{tabular}{c}
     PNG DHS \\ 2016–18 \end{tabular} &	 3,380 	& \checkmark&	N/A	&N/A&	N/A	&N/A	& \checkmark& \checkmark & \checkmark& \checkmark&	N/A	& \checkmark&	N/A\\
\hline
Bitew et al.~\cite{ethiopia} &	LR, RF, KNN,  ANN, \textbf{XGBoost}	& Ethiopia & 2021	&	EDHS 2016 & 9,471 &	\checkmark&	N/A	& N/A	&N/A &	N/A	&\checkmark&	N/A	&\checkmark&	N/A	& N/A &	N/A	&\checkmark\\
\hline
 Jain et al.~\cite{india}	&LR, RF, ANN, XGBoost, \textbf{TabNet}, Pytorch Tabular, ANN, NB, \textbf{AutoML} &	India	&2022	& \begin{tabular}{c} IDHS 2005-06 \\ 2015-16 \end{tabular} &	- &	\checkmark &	\checkmark	&N/A &	\checkmark &	N/A	&\checkmark &	N/A	& N/A	& N/A	& N/A &	\checkmark &	N/A \\
\hline
Rahman et al.~\cite{bangladesh3} &	LR, SVM, \textbf{RF}	& Bangladesh & 2021 & BDHS 2014 &	 7,079 &	\checkmark &	\checkmark	& \checkmark	& N/A	& N/A	& \checkmark &	N/A &	N/A &	N/A &	N/A &	\checkmark & N/A \\
\hline
 Van et al.~\cite{philippines} &	\textbf{RF}, SVM, LDA, LR & Philippines	& 2021 &	Primary Source &	 618 &	AMDR & NAM & PDRI &	N/A & N/A &				\checkmark&	N/A	&\checkmark&	N/A&	N/A	&N/A	&N/A \\
\hline
	Fenta et al.~\cite{ethiopia2}	& LR, Ridge,  Lasso, ANN,   Elastic-Net, \textbf{RF} &	Ethiopia	& 2021	& EDHS &	29,333	& N/A	& N/A	& N/A &	N/A &	\checkmark	& \checkmark &	\checkmark &	\checkmark &	\checkmark &	N/A	& \checkmark &	N/A\\
\hline
Talukder et al.~\cite{bangladesh1} &	LDA, KNN,SVM, \textbf{RF}, LR &	Bangladesh	&2020 &	BDHS 2014 &	 6,863 &	N/A	&N/A	&\checkmark	&N/A	&N/A	&\checkmark&	N/A&	\checkmark&	N/A&	N/A	&N/A&	\checkmark\\
\hline
Mani et al.~\cite{usa} &	Multi LR, LDA, \textbf{RF} &	USA &	2018 & NHANES &	145,263	 & N/A	& N/A &	\checkmark	& N/A &	N/A &	\checkmark &	\checkmark &	\checkmark &	\checkmark &	N/A	& N/A &	N/A \\\hline
\end{tabular}
}}}
\captionsetup{hypcap=false} 
\captionof{table}{ Comparative summary across existing literature}
\label{tab:previous_studies}
\end{minipage}%
\hfill
\normalsize
\begin{minipage}[t]{0.39\textwidth}
Table~\ref{tab:previous_studies} summarizes how our work compares with existing international studies. Like \cite{india}, we identified TabNet as a top performer; however, their study omitted weight-for-age, a key anthropometric indicator, and selected models solely by accuracy and AUC, potentially neglecting class imbalance. Our methods are similar to \cite{ethiopia2} in conducting a broad model comparison, but their exclusion of TabNet, SVM, and advanced gradient boosting led to different conclusions about the best models. Several studies, including \cite{egypt}, \cite{bangladesh1}, \cite{papua_new_guinea}, and \cite{ethiopia}, identified XGBoost as the top classifier, though from limited model sets that did not include TabNet. In contrast, work from Bangladesh \cite{bangladesh3}, the Philippines \cite{philippines}, Ethiopia \cite{ethiopia2}, and the USA \cite{usa} favored random forest, again without TabNet and often without XGBoost. Tamanna et al. \cite{first} also reported strong random forest performance but without cross-validation or hyperparameter tuning, likely inflating performance estimates.

An Ethiopian study \cite{ethiopia} (n = 9,471) achieved accuracy 0.67, sensitivity 0.71, and recall 0.57 with XGBoost, while another Ethiopian study \cite{ethiopia2} (n = 29,333) reported accuracy 0.67, sensitivity 0.52, specificity 0.81, precision 0.73, and F1-score 0.61. Research in Papua New Guinea using XGBoost \cite{papua_new_guinea} obtained F1-score 0.66, recall 0.62, accuracy 0.72, and precision 0.71. Outcome definitions vary substantially across studies. Some studies focus on single anthropometric indicators (stunting only in \cite{papua_new_guinea, ethiopia}), others examine all three indicators separately (stunting, wasting, and underweight as distinct outcomes), and \cite{ethiopia2} use composite indicators similar to our approach as documented in Table ~\ref{tab:previous_studies}. A Philippine study reported the highest accuracy (0.78) and sensitivity (0.90) with random forest \cite{philippines}, but its very low specificity (0.16) indicates substantial overprediction of malnutrition, underscoring the need for balanced metrics. Such imbalance, misclassifying nourished children as malnourished, can cause unnecessary clinical workload and resource misallocation. Our TabNet model attains comparable or better F1-score (0.62) with balanced precision (0.63) and recall (0.62), avoiding the extreme trade-offs seen in some prior studies. \\[1ex]
\end{minipage}

\subsection*{\small Recommendation}
Given the modest predictive performance achieved across all models, the following recommendations should be considered as research-informed suggestions rather than definitive policy directives. This study suggests that the proposed TabNet deep learning model may offer marginal advantages over widely used traditional algorithms such as RF or XGBoost for exploratory prediction of malnutrition in survey-based contexts, though further validation is needed before operational deployment. Our findings provide supportive evidence for existing evidence-based interventions rather than novel policy directions. Poor households could benefit from focused support to ensure adequate diets for their children, as the wealth index consistently emerged as a key predictor of nutritional status in our analysis. Women's education appears to be an important factor and warrants continued prioritization in efforts to reduce child malnutrition. The model-identified determinants for the Nepali context may offer useful insights for public health professionals and policy makers working to reduce malnutrition, complementing existing knowledge from epidemiological studies. The observed effect of spatial attributes suggests potential value in considering differential resource allocation across provinces.

\subsection*{\small Limitation and Future Work}
The NMICS dataset relies on self-reported data, introducing recall and social desirability bias. Although extensive for a demographic survey, the feature set is relatively limited for complex predictive modeling, and substantial missing data in some variables required imputation, which may have affected model performance. The moderate predictive accuracy likely stems from the indirect nature of sociodemographic predictors and the relatively small sample size ($n=6,416$) for training machine and deep learning models. Future work should incorporate other potential factors such as maternal nutritional status and health during pregnancy, gestational age, breastfeeding practices and duration, and micronutrient supplementation to achieve better predictive performance. Future research should explore ensemble approaches combining TabNet's attention mechanism with SVM's discriminative boundaries, implement cost-sensitive learning with asymmetric penalties for false negatives, and investigate transfer learning from larger multi-country survey datasets to improve performance on Nepal-specific data. Combining stunting, wasting, and underweight into a single binary outcome obscures clinically important distinctions between underweight, stunting, and wasting, which require different intervention strategies. Our composite approach suits population-level screening but limits clinical actionability for intervention type selection. Future research should develop condition-specific prediction models to enable more targeted clinical decisions.

\section*{Conclusions}

Understanding childhood malnutrition through predictive modeling can guide evidence-based policies and targeted interventions from household counseling to national nutrition programs. Deep learning models, especially TabNet, achieved the highest scores among gradient boosting and traditional machine learning. After hyperparameter tuning, TabNet demonstrate top-ranked model based on combined metric performance performance (F1: 0.62, recall: 0.62, precision: 0.63), using sequential attention to identify key features at each decision step. Among traditional models, support vector machine performed best (F1: 0.61, ROC-AUC: 0.64), and AdaBoost led gradient boosting methods (F1: 0.61, average precision: 0.56). Child age, maternal education, household wealth, meal frequency, and geographic region consistently emerged as key predictors, highlighting the socioeconomic and demographic drivers of malnutrition. These results support targeted preventive interventions and better allocation of health resources in Nepal. The identification of high-risk profiles, children in specific age groups, from poorer households, with less-educated mothers, and in vulnerable provinces, can inform focused screening and early intervention. By enabling early detection and risk stratification, these models can help Nepal advance toward SDG 2 (zero hunger) and SDG 3 (good health and well-being) by reducing childhood malnutrition.

The study’s framework and findings are applicable to other low- and middle-income countries with similar demographic and socioeconomic contexts, especially those using survey data. Future work should incorporate clinical and dietary variables, test ensembles of top-performing models, and assess real-world implementation in resource-limited health systems to maximize public health impact.

\section*{Code and Dataset Availability}
Python programming language was used for this work and the datasets used in this study are publicly available from the Multiple Indicator Cluster Survey repository (https://mics.unicef.org/surveys). All author-generated code for data preprocessing, feature selection, model training, evaluation, and visualization is publicly available on GitHub (\href{https://github.com/deepakbastola/malnutrition}{https://github.com/deepakbastola/malnutrition}) under the MIT License, ensuring unrestricted access for reproducibility and reuse.

\nolinenumbers

\bibliography{ref}

\newpage
\section*{Appendix}
 
\subsection*{Separability}

Following Fig~\ref{fig8} principal component analysis (PCA), t-distributed stochastic neighbor embedding (t-SNE), and uniform manifold approximation and projection (UMAP) shows that the features used in the model are not fully sufficient to easily distinguish between nourished and malnourished individuals. Therefore, we employed sophisticated machine learning and deep learning algorithms.

\begin{figure}[!ht]
    \centering
     \includegraphics[width=0.75\linewidth]{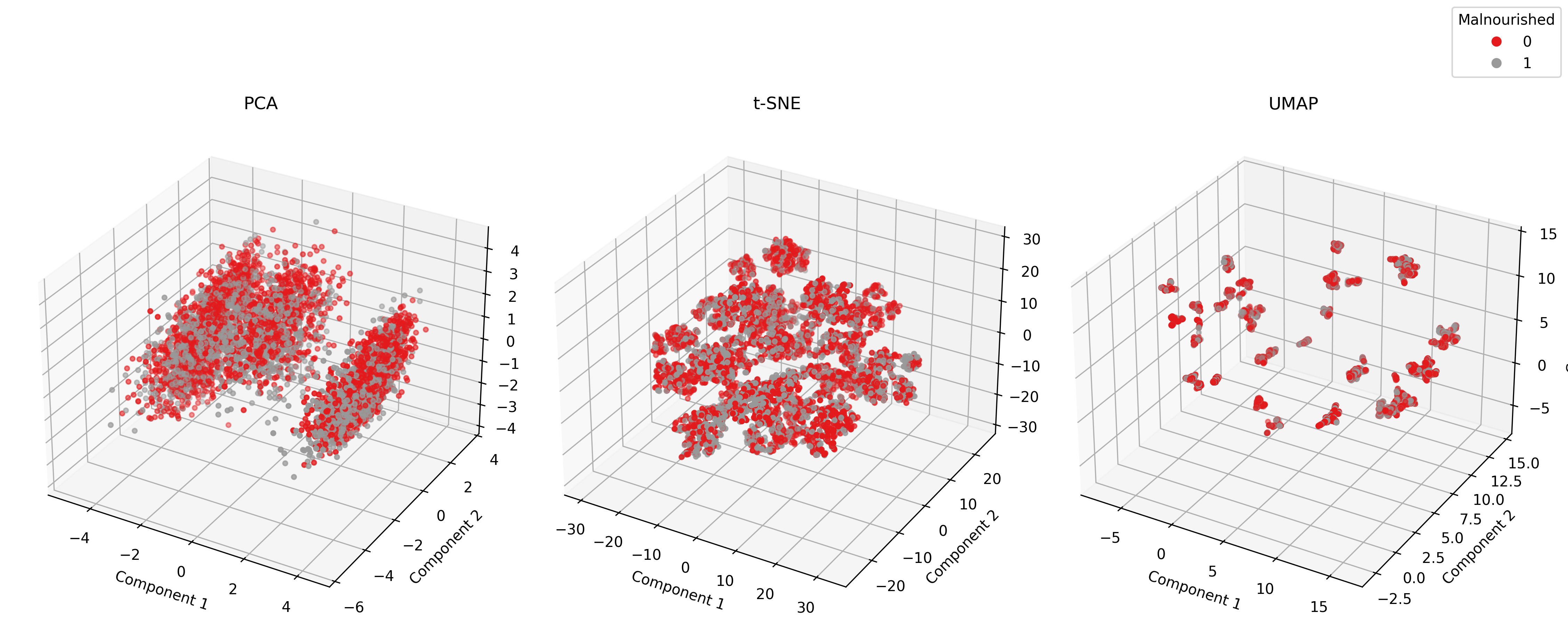}
    \caption{\bf Class separability using PCA, t-SNE and UMAP}
    \label{fig8}
\end{figure}

\subsection*{Performance Profile}

    \begin{minipage}{0.53\textwidth}
    \raggedright
    Fig~\ref{fig9} presents a comparative performance profile across model categories using six key metrics. Deep learning models (blue solid line) demonstrate the largest coverage area, achieving the highest scores in recall (0.61), precision (0.61), F1-score (0.61), and balanced accuracy (0.61), while matching other categories in ROC-AUC (0.63). Traditional machine learning methods (green dotted line) and gradient boosting algorithms (orange dashed line) exhibit nearly overlapping profiles, with both reaching ROC-AUC of 0.63-0.64 but showing slightly lower performance in recall and F1-score (approximately 0.60). All three categories converge at similar accuracy levels (~0.61), indicating that the performance advantage of deep learning manifests primarily in precision-recall balance and class-specific metrics rather than overall accuracy. The symmetric radar pattern for gradient boosting and traditional machine learning suggests comparable strengths across all evaluation dimensions, while deep learning's expanded polygon reflects more pronounced capability in handling the imbalanced malnutrition classification task.
    \end{minipage}
    \hfill
   \begin{minipage}{0.46\textwidth}
    \centering
    \includegraphics[width=\linewidth]{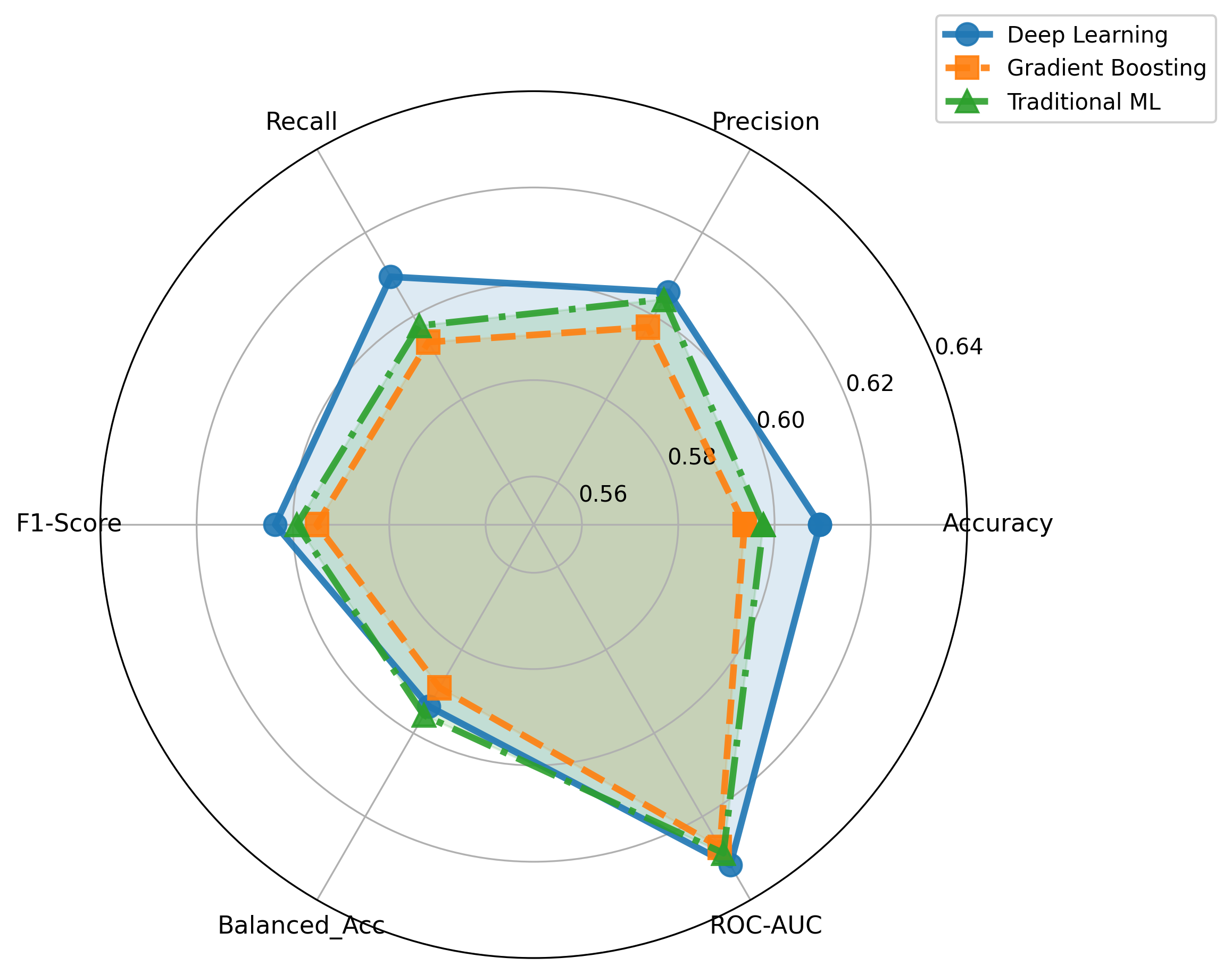}
    \captionof{figure}{\textbf{Performance profile of the top three models}}
    \label{fig9}
\end{minipage}

\subsection*{Hyperparameter and Reproducibility Implementation}
To ensure full reproducibility of all reported results, the following measures together with hyperparameter (presented in Table~\ref{tab:all_hyperparameters}) were implemented:

\paragraph{Random Seed Control:} A global seed value of 42 was applied across all stochastic components, including Python's built-in \texttt{random} module, NumPy, TensorFlow, and PyTorch. This was enforced through \texttt{PYTHONHASHSEED=42}, \texttt{random.seed(42)}, \texttt{np.random.seed(42)}, \texttt{tf.random.set\_seed(42)}, \\
\texttt{tf.keras.utils.set\_random\_seed(42)}, \texttt{torch.manual\_seed(42)}, and \texttt{torch.cuda.manual\_seed\_all(42)}. Individual Keras models employed distinct seeds (DNN: 42, ResNet: 43, Wide \& Deep: 44) to prevent identical weight initialization across models while maintaining within-model reproducibility.
\paragraph{Deterministic Operations:} Environment variables \texttt{TF\_DETERMINISTIC\_OPS=1} and \texttt{TF\_CUDNN\_DETERMINISTIC=1} were set to enforce deterministic GPU operations in TensorFlow. PyTorch determinism was configured through \texttt{torch.backends.cudnn.deterministic=True} and \texttt{torch.backends.cudnn.benchmark=False}, eliminating non-deterministic algorithmic choices.
\paragraph{Reproducible Data Processing:} The train-test split (80:20 ratio) used \texttt{random\_state=42} with stratified sampling to preserve class proportions across partitions. SMOTE oversampling was applied with \texttt{random\_state=42} to generate consistent synthetic minority class samples within the training data.
\paragraph{Explicit Weight Initialization:} All neural network layers used explicitly seeded initializers -- Glorot Uniform for sigmoid and linear output layers, and He Normal for ReLU-based hidden layers in the Wide \& Deep architecture. Per-layer seed offsets prevented identical weights across layers within the same model.

\begin{table}[!ht]
\centering
\scriptsize
\resizebox{\textwidth}{!}{

\begin{tabular}{llp{8cm}}
\toprule
\textbf{Category} & \textbf{Model} & \textbf{Hyperparameters} \\
\midrule
\multirow{7}{*}{\rotatebox{90}{\textbf{Traditional ML}}} 
& Logistic Regression & max\_iter=1000, random\_state=42, default L2 regularization \\
& KNN & n\_neighbors=5 \\
& LDA & default solver (svd) \\
& Decision Tree & max\_depth=10, random\_state=42 \\
& Random Forest & n\_estimators=200, max\_depth=15, min\_samples\_split=5, n\_jobs=-1, random\_state=42 \\
& Extra Trees & n\_estimators=300, max\_depth=20, min\_samples\_split=4, min\_samples\_leaf=2, max\_features='sqrt', bootstrap=True, n\_jobs=-1, random\_state=42 \\
& SVM & kernel='rbf', probability=True, random\_state=42 \\
\midrule
\multirow{5}{*}{\rotatebox{90}{\textbf{Gradient Boosting}}} 
& XGBoost & n\_estimators=200, max\_depth=6, learning\_rate=0.05, subsample=0.8, colsample\_bytree=0.8, eval\_metric='logloss', n\_jobs=-1, random\_state=42 \\
& LightGBM & n\_estimators=300, max\_depth=7, learning\_rate=0.05, num\_leaves=50, subsample=0.8, colsample\_bytree=0.8, min\_child\_samples=20, reg\_alpha=0.1, reg\_lambda=0.1, n\_jobs=-1, random\_state=42 \\
& CatBoost & iterations=200, depth=6, learning\_rate=0.05, l2\_leaf\_reg=3, subsample=0.8, random\_state=42 \\
& Hist Gradient Boosting & max\_iter=200, max\_depth=7, learning\_rate=0.05, l2\_regularization=1.0, random\_state=42 \\
& AdaBoost & base\_estimator=DecisionTree(max\_depth=3), n\_estimators=200, learning\_rate=0.1, random\_state=42 \\
\midrule
\multirow{4}{*}{\rotatebox{90}{\textbf{Deep Learning}}} 
& DNN (Keras) & Architecture: 256→128→64→1; Activation: LeakyReLU(0.1); Optimizer: Adam (lr=0.001); Dropout: 0.3, 0.3, 0.2; L2: 0.001; BatchNorm: yes; Loss: binary\_crossentropy; Batch=64; Epochs=100 (patience=20); Initializer: GlorotUniform \\
& ResNet (Keras) & Architecture: 128→ResBlock(128)→ResBlock(64)→1; Activation: LeakyReLU(0.1); Optimizer: Adam (lr=0.0005); Dropout: 0.3, 0.2; L2: 0.001; BatchNorm: yes; Loss: binary\_crossentropy; Batch=64; Epochs=100 (patience=20); Initializer: GlorotUniform \\
& Wide \& Deep (Keras) & Wide: 128→64 (ReLU); Deep: 384→192→96 (LeakyReLU); Combined: 128→64→32→1; Optimizer: Adam (lr=0.001); Dropout: 0.25-0.40; L2: 0.0005-0.001; BatchNorm: yes; Loss: binary\_crossentropy; Batch=64; Epochs=100 (patience=20); Initializer: HeNormal \\
& TabNet (PyTorch) & n\_d=64, n\_a=64, n\_steps=5, gamma=1.5, lambda\_sparse=1e-4, momentum=0.02, clip\_value=2.0; Optimizer: Adam (lr=0.02, weight\_decay=1e-5); Scheduler: StepLR (step\_size=50, gamma=0.95); mask\_type='sparsemax'; Batch=1024 (virtual=256); Epochs=200 (patience=50) \\
\bottomrule
\end{tabular}
}
\caption{Complete hyperparameter specifications for all 16 machine learning and deep learning models}
\label{tab:all_hyperparameters}
\end{table}

\subsection*{Feature Selection Ablation Analysis}
\label{appendix:ablation_analysis}
Table~\ref{tab:ablation_analysis} presents the comparative performance of TabNet for strict Boruta-confirmed vs. multi-criteria consensus feature sets. Both feature sets were evaluated using identical preprocessing pipelines, train-test splits (random\_state=42), SMOTE oversampling on training data only, and TabNet hyperparameter configurations to ensure fair comparison.

\begin{table}[!ht]
\centering
\begin{tabular}{lccc}
\toprule
\textbf{Metric} & \textbf{Boruta-Confirmed} & \textbf{Multi-Criteria} & \textbf{Difference} \\
\midrule
Accuracy & 0.60 & 0.62 & +0.02 \\
Precision & 0.60 & 0.63 & +0.03 \\
Recall & 0.60 & 0.62 & +0.02 \\
F1-Score & 0.60 & 0.62 & +0.02 \\
ROC-AUC & 0.62 & 0.64 & +0.02 \\
Average Precision & 0.53 & 0.54 & +0.01 \\
Balanced Accuracy & 0.59 & 0.62 & +0.03 \\
Cohen's Kappa & 0.19 & 0.24 & +0.05 \\
MCC & 0.19 & 0.24 & +0.05 \\
Brier Score & 0.24 & 0.24 & 0.00 \\
\bottomrule
\end{tabular}
\caption{TabNet Performance Comparison}
\label{tab:ablation_analysis}

\end{table}

The multi-criteria consensus approach demonstrated consistent improvements over strict Boruta-only selection across all evaluation metrics. The higher performance of the multi-criteria approach validates our methodological choice to retain features that demonstrated strong support across multiple ensemble ranking methods and established domain relevance.

\end{document}